\documentclass{ar-1col-S2O}
\usepackage[numbers]{natbib}
\usepackage{url}
\usepackage{todonotes}
\usepackage{bm}
\usepackage{amsmath}
\usepackage{wrapfig}    % probably to be removed as we are required to only use the figure package
\usepackage{comment}
\usepackage{multirow}
\usepackage{rotating}  % To rotate the table

\usepackage{xcolor,colortbl}
\definecolor{lavender}{rgb}{0.9, 0.9, 0.98}
\definecolor{red}{RGB}{255, 0, 0}
\definecolor{orange}{RGB}{252, 130, 62}
\definecolor{blue}{RGB}{0, 0,255}
\definecolor{darkgreen}{RGB}{0, 150,0}
\newcommand{\note}[1]{\color{orange}}
\usepackage{booktabs}

\newcommand{\rxmark}{\textcolor{red}{\xmark}}
\newcommand{\gcmark}{\textcolor{green}{\cmark}}

\usepackage{pifont}% http://ctan.org/pkg/pifont
\newcommand{\cmark}{\ding{51}}%
\newcommand{\xmark}{\ding{55}}%
\usepackage{booktabs}

\usepackage[nolist]{acronym} % acronyms by the \ac{label} command
\newacro{rl}[RL]{Reinforcement Learning}
\newacro{il}[IL]{Imitation Learning}
\newacro{dr}[DR]{Domain Randomization}

\jname{Xxxx. Xxx. Xxx. Xxx.}
\jvol{AA}
\jyear{YYYY}
\doi{10.1146/((please add article doi))}

\begin{document}

% Page header
\markboth{Longhini et al.}{A Review of Robotic Cloth Manipulation}

% Title
\title{Unfolding the Literature: A Review of Robotic Cloth Manipulation}

%Authors, affiliations address.
\author{Alberta Longhini$^1$, Yufei Wang$^2$, Irene Garcia-Camacho$^3$, David Blanco-Mulero$^3$,  Marco Moletta$^1$, Michael Welle$^1$, Guillem Alenyà$^3$, Hang Yin$^4$, Zackory Erickson$^2$, David Held$^2$, Júlia Borràs$^3$, Danica Kragic$^1$ 
\affil{$^1$Department of Robotics, Perception, and Learning, KTH Royal Institute of Technology,
100 44 Stockholm, Sweden; email: albertal@kth.se, marco.moletta@gmail.com, mwelle@kth.se, dani@kth.se }
\affil{$^2$The Robotics Institute, Carnegie Mellon University, Pittsburgh, USA, PA 15213; email: yufeiw2@andrew.cmu.edu, zerickso@andrew.cmu.edu
, dheld@andrew.cmu.edu }
\affil{$^3$Institut de Robòtica i Informàtica Industrial, CSIC-UPC, Barcelona, Spain, 08028; email: igarcia@iri.upc.edu, dblancom@iri.upc.edu, galenya@iri.upc.edu, jborras@iri.upc.edu}
\affil{$^4$Department of Computer Science, University of Copenhagen, Copenhagen, Denmark, 2100; email: hayi@di.ku.dk}}

%Abstract
\begin{abstract}

The realm of textiles spans clothing, households, healthcare, sports, and industrial applications. The deformable nature of these objects poses unique challenges that prior work on rigid objects cannot fully address. The increasing interest within the community in textile perception and manipulation has led to new methods that aim to address challenges in modeling, perception, and control, resulting in significant progress. However, this progress is often tailored to one specific textile or a subcategory of these textiles. 
To understand what restricts these methods and hinders current approaches from generalizing to a broader range of real-world textiles, this review provides an overview of the field, focusing specifically on how and to what extent textile variations are addressed in modeling, perception, benchmarking, and manipulation of textiles.
%To understand what restricts these methods and hinders current approaches from generalizing to a broader range of real-world textiles, this review provides an overview of the field.  Specific focusing on how, and to what extend, textiles variations are addressed from the modeling, perception and control perspectives. %We examine existing approaches for modeling, perception, control, and learning architectures to manipulate textile objects. 
%We further summarize how recent works address generalization %\dbm{and adaptation, although this might be missleading without introducing the term} 
%to textile variations, investigating their potential and limitations. 
We finally conclude by identifying key open problems and outlining grand challenges that will drive future advancements in the field.
%Finally, we outline open problems and grand challenges to advance the field.%Finally, we offer future perspectives on key challenges to advance the field\dbm{This is too generic, I would prefer some end sentence about the grand challenges or the open problems}.
\end{abstract}

%Keywords, etc.
\begin{keywords}
Textiles, Deformable Object Manipulation, Generalization,  Physical Properties Variations, Tasks Variations
\end{keywords}
\maketitle

%Table of Contents
% \tableofcontents

% \include{sections/01_Introduction}
% \include{sections/02_Background}
% \include{sections/03_Modeling}
% \include{sections/04_Perception}
% \include{sections/05_Manipulation}
% % \include{sections/old_versions/05_Manipulation}
% \include{sections/06_0_Synergy}
% \include{sections/06_Resources}
% \include{sections/07_Applications}
% \include{sections/08_Conclusions}

% Heading 1
\section{INTRODUCTION}

%\todoinb{Alberta - Integrate Citations}
% \todoinor{Alberta - Shrink to 1 page}
% \todoinor{Alberta - Address Feedback}

% Motivation and current challenges
Textile deformable objects such as clothing items or household objects like bed sheets and blankets are ubiquitous in our daily lives. Their usage spans applications from healthcare and domestic environments to the textile industry. 
Efforts to automate the manipulation and processing of these objects promise to enhance recycling and textile reuse while providing greater assistance to aging populations. Despite recent advances in manipulation tasks such as assistive dressing~\cite{kapusta2016data, chance2017quantitative}, folding or bagging~\cite{seita2021learning, lippi2020latent}, textile manipulation remains challenging as the deformable nature of these objects breaks fundamental assumptions in robotics such as rigidity, known dynamics models, and low dimensional state space~\cite{zhu2022challenges}.  Specifically, when forces are applied to a deformable body, they not only move the object but also change its shape. From a physics point of view, understanding how the shape changes requires knowledge about the object's physical properties, such as stiffness or elasticity. From a perceptual point of view, properties such as shape, color, and material provide distinct signals to visual and tactile sensors.  Endowing robots with skills to perceive, manipulate, and address the diversity of these textile properties presents a compelling avenue toward autonomous agents. However, current methods for textile manipulation tend to be tailored to specific objects that mirror the properties of the simulators used in their design~\citep{lippi2020latent,huang2023self,matas2018sim, salhotra2022learning,ma2022learning}. 

This review consolidates recent methods and applications of deformable object manipulation, specifically textiles, highlighting the challenges associated with variations in their physical properties. We evaluate the progress made and identify key areas requiring further research to enhance the generalization and adaptive capabilities of robots in handling real-world textiles.  Recent reviews have explored specific methods and applications related to this object category. Particular emphasis has been given to grasping~\cite{borras2020grasping} and caregiving scenarios ~\cite{jimenez2020perception,wang2023deformable}. In contrast, our review broadens the scope by analyzing generalization across different textile variations and applications.  Building on the foundational work on perception, modeling, and manipulation of deformable objects~\cite{arriola2020modeling,yin2021modeling},  our work offers a distinct perspective on enhancing the generalization and adaptability of perceptual and manipulation skills.  Similarly to~\cite{zhu2022challenges}, we seek to identify ongoing challenges in modeling, perception, and control, further addressing the underexplored area of %generalization and 
benchmarking.% specifically for textile objects.
The review is organized as follows. In Section~\ref{sec:background}, we provide the fundamentals about textiles, covering the variations of properties and tasks that will be discussed throughout the document. Section~\ref{sec:modeling} reviews analytical and learning methods to model textile dynamics, highlighting their connection to textile physical properties. Section~\ref{sec:perception} identifies current approaches to perceive textile properties,  whereas Section~\ref{sec:manipulation} covers current approaches for textile manipulations with a focus on techniques that enable generalization and adaptability to variations of properties and to what extent do current methods account for variations in physical and mechanical properties and tasks.
% In Section~\ref{sec:synergies}, we provide the interplay between modeling, perception, and manipulation, while i
In Section~\ref{sec:resources}, we discuss currently available resources such as benchmarks and datasets, that enables evaluating this generalization, including benefits and limitations. 
We provide a thorough overview of application areas in Section~\ref{sec:application}. We close with a discussion about the interplay between modeling, perception, and manipulation, and future perspectives in Section~\ref{sec:conclusions}.

\section{FUNDAMENTALS}
\label{sec:background}

This section provides an overview of the fundamental aspects of textiles, detailing the definitions and characterizations of textile objects, exploring the variations in their physical properties, and examining the diversity in manipulation tasks. Understanding these foundational elements is essential for delineating the types of variations and challenges discussed in the subsequent sections on modeling, perception, and manipulation. 

\subsection{Textiles, Fabrics, and Cloths}
\label{sec:background:textiles}
% \begin{figure}
%   \centering
%   \includegraphics[width=0.48\textwidth]{images/02_PyramidComplex.png} 
%   \caption{\al{make the image more aesthetically beautiful} \ig{We could include as well the concepts used in the text (fabric and cloth)} The textile production process affects the physical properties of deformable objects, such as elasticity, stiffness, and friction. The further an object goes through the production process, the more complex properties like elasticity, surface friction, and stiffness become. We can identify three steps of the production process: a) the starting point is the choice of the material of the thread; b) threads are subsequently intertwined to form textiles; c) textiles are finally sewn together to originate specific types of deformable objects (e.g., garments, household objects, or bandages). \dave{what are bed clothes?} \dave{This figure is confusing, where the y-axis is both ``increasing complexity" and also ``material, construction technique, and type"}}
%   \label{fig:CDOs_pyramid}
% \end{figure}

\begin{figure}
  \centering
  \includegraphics[width=1.\textwidth]{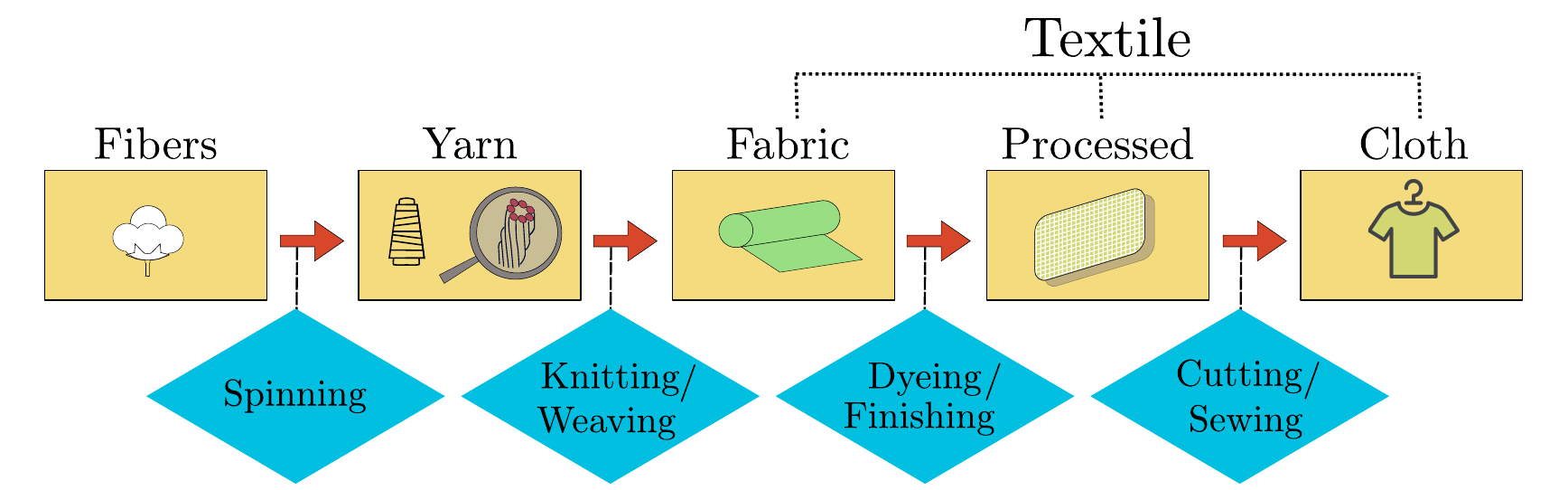} 
  \caption{\textbf{Manufacturing process of textiles:} textile is an umbrella term that covers materials that are made of interlacing natural or synthetic fibers. The figure depicts the textile production process in its different stages, where yellow boxes represent materials and objects, while blue blocks specify the processing step. }
  \label{fig:prod_process}
\end{figure}

\begin{marginnote}[]
\entry{Textile}{originates from the Latin \textit{textilis}, meaning 'woven' and derived from the verb to weave.}
\end{marginnote}

Textile has evolved from its initial reference, woven fabrics, to encompass a broad spectrum of objects. 
These include traditional woven fabrics, as well as deformable and flexible materials made from yarns or threads through various construction processes beyond weaving. Common terms used to describe textile objects are \textbf{fabric} and \textbf{cloth}. Despite being often overlapped in usage, they carry subtle distinctions tied to different phases of the production process, as shown in Fig.~\ref{fig:prod_process}. Fabric is any thin, flexible material crafted from thread or yarn, fibers, polymeric film, foam, or their combinations, and is used in creating further products like clothing, requiring additional production steps \cite{Kadolph2007Textiles}. Cloth, while sometimes used interchangeably with fabric, typically refers to fabric that has undergone further processing or cutting. Everyday clothing items are predominantly made through weaving or knitting before being sewn, with high fashion exploring other methods more extensively. 

The physical and mechanical properties of the final textile object are intrinsically tied to its manufacturing process. Yarns and threads, the building blocks of these textiles, are spun from various origins—animal, plant, mineral, synthetic, or blends thereof.  Woven fabrics, produced by interlacing two sets of threads, are known for their hardness and non-elasticity, making them ideal for garments like shirts and jeans that retain creases. In contrast, knitted fabrics, created by interloping a single set of yarn, offer softness and elasticity in all directions and thus are used in wrinkle-resistant clothing such as t-shirts, which stretch uniformly in all directions to better conform to the body's shape. Additionally, further production steps involve cutting and coloring fabrics to create clothing with various shapes and colored textures. A comprehensive exploration of fabric properties is given in~\cite{grishanov2011structure}.

\subsection{Variations of Object Properties}
\label{subsec:Var_Properties}

%\TODO{Discuss the purpose and structure of this section.}

% \al{To brainstorm and to keep in mind:
% \begin{itemize}
%     \item Do we want to provide more in-depth details about these properties? How do they connect to the properties used for modeling cloths?
%     \item Do we want to have definition blocks for each of these properties?
%     \item How do we integrate the spider plot?
% \end{itemize}
% }

\begin{figure}
  \centering
  \includegraphics[width=\textwidth]{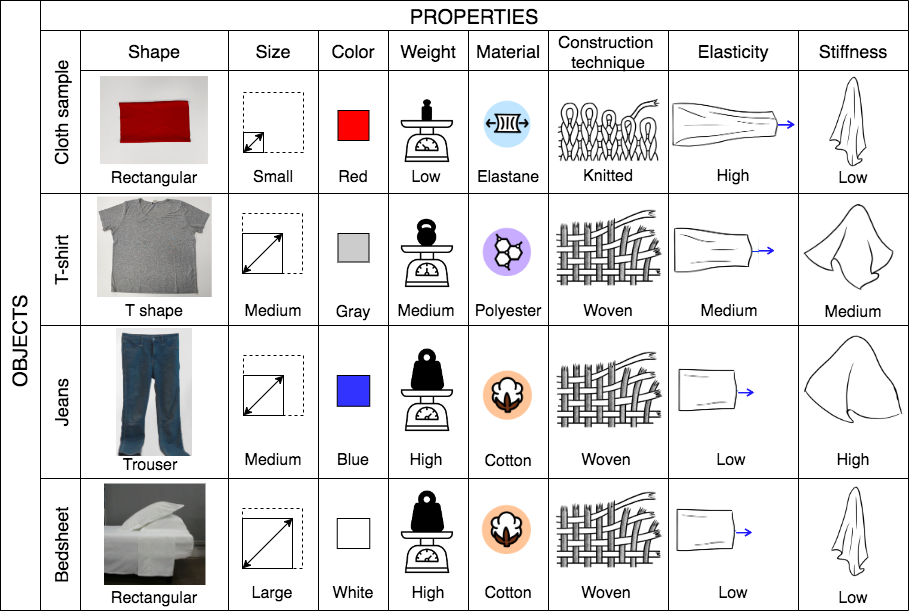}
        \caption{\textbf{Object properties variations: } Object differences rely on the variation of their physical and mechanical properties. The figure shows visual examples of how properties can vary between four objects from literature cloth sets. Measures can be found in \cite{garcia-camacho2024standardization}.}% \cite{longhini2023edonet, Gustavsson2022Cloth, garcia-camacho2022hcos}.}%; A cloth sample, a T-shirt, jeans and a bedsheet.}
        \label{fig:prop_variations}
        \vspace{-3pt}
\end{figure}

One of the major challenges in robotic manipulation is designing planning and control algorithms that generalize or adapt to variations that will be encountered in real-world deployments~\cite{cui2021toward}. In this review, we refer to \textbf{generalization} as the ability to apply the knowledge acquired from a specific set of textile properties to unseen variations of these properties. \textbf{Adaptation} is instead the ability to dynamically adjust strategies or models in response to changing conditions or novel variations in the object properties. Given the challenges of generalization and adaptation, it is crucial to first understand the variations in textile properties, which can be broadly classified into physical and mechanical properties.  Fig.~\ref{fig:prop_variations} shows an example of different variations of textile properties for four different objects.

\subsubsection{Physical properties}
\label{sec:physical_properties}

% \begin{textbox}[]
% \section{Physical Properties}
% Sidebar text goes here.
% \subsection{Color}
% More text goes here.
% \subsubsection{Sidebar third-level heading}
% Text goes here.
% \end{textbox}

The object's physical properties are inherent characteristics and can be observed and measured without subjecting the material to external forces or manipulations. These properties describe how a material appears under static or non-changing conditions. Among these properties, 
\textbf{Size} and \textbf{Shape} are crucial as they influence the robot's workspace, correlating with the object's function and category. %The \textbf{Weight} is one of the first properties that affect the manipulation of these objects. 
The \textbf{Weight} influences how the textile deforms under gravity. For heavy and large textiles such as blankets, %the weight might be important depending on the maximum payload that a robotic manipulator can lift. Big heavy items
robots may need bimanual systems to manipulate them. The \textbf{Color} affects perception-based algorithms, essential for identifying and tracking the textile during manipulation tasks~\cite{kapusta2016data}. \textbf{Fabric Material} composition, ranging from natural to synthetic fibers, dictates interaction forces between the textile and the robot, altering manipulation strategies~\cite{longhini2021textile}. The \textbf{Construction Technique}, referring to the knitting or weaving processes, directly impacts a fabric's mechanical behavior by defining, for example, how tight or loose the construction pattern is, thus influencing the object elasticity and rigidity~\cite{longhini2023elastic}.
%such as elasticity, which can be observed through the distinct behavior of woven versus knitted fabrics under stress~\cite{longhini2023elastic}. %\dave{I am not sure that construction technique is directly relevant, except in how it affects the mechanical properties described in the next section.  Same with material.  Overall I am thinking that we should probably eliminate this section and briefly mention material and construction technique in the context of the mechanical properties in the next section.} \dave{If we keep this section, can we modify it to talk more about the papers that address these variations?}

\subsubsection{Mechanical properties}
%\TODO{Decide how to integrate the following comment from DMB}
%\dbm{One common way of defining the dynamics of an object is by its strain/stress curve, the Poisson's ratio and the Young's modulus. These parameters are available in some simulation engines (MuJoCo and Isaac Sim), so it might be good either to mention them here, or as future work identifying these together with the mechanical properties that are already discussed in the text}
%\hy{after reading the texts below, I also feel it would be a good opportunity to introduce Young's modulus and Poisson's ratio here.}

The mechanical properties are parameters that describe how a material responds to applied forces or manipulations. Cloth \textbf{Stiffness} or rigidity influences how it behaves under manipulation as it determines the resistance to deformation.  %Understanding cloth stiffness can help in predicting cloth behavior when manipulating and, therefore, improve planning tasks.  
Stiffness is a key determinant of a fabric's manipulation behavior, as it quantifies the textile's resistance to bending and deformation. %Accurately reasoning about stiffness is crucial for predicting how a textile will behave when handled by a robot, thus facilitating improved planning and execution of manipulation tasks~\cite{sanchez2018robotic}.  
%The stiffness is traditionally measured using principles derived from the Cusick drape test~\cite{cusick1968}, which compares the draped area of the cloth to its original, undraped state. 
\textbf{Elasticity}, or the capacity of a textile to stretch and recover to its original size after being deformed, is essential, particularly in applications like robotic dressing assistance~\cite{sun2024force}. High elasticity can accommodate user movements and minimize safety risks during interaction. %The elasticity measurement is based on the extent to which the fabric stretches from its original length when force is applied. %Further to the traditional measures of elasticity, understanding the broader dynamics of a textile's response to force, such as its strain/stress curve, can provide deeper insights into its behavior \al{cite}. These dynamics are quantitatively described by parameters like the Young's modulus, which measures the stiffness of an elastic material, and the Poisson's ratio, which describes the ratio of the transverse strain to the axial strain in a stretched material. 
% \dbm{Should you discussed in more detail Poisson's and Young's modulus? These are also the parameters that are usually found in simulation engines. Check comments.}
Elasticity and stiffness are properties of the material that are often characterized by Young’s modulus and Poisson’s ratio. The \textbf{Young's modulus} quantifies a material's ability to resist deformation under stress by showing the relation between stress and strain in the elastic region of the textile. The \textbf{Poisson's ratio} measures instead the ratio of transverse strain to axial strain in a stretched material. %Finally, \textbf{Friction} is the force resisting the relative motion of solid surfaces, fluid layers, and material elements sliding against each other. 
Finally, \textbf{Friction} refers to the resistance encountered when one surface slides over another. In the context of textiles, friction can vary significantly depending on the surface characteristics of the fabric~\cite{li2015folding}. Friction properties affect how textiles move against surfaces, thus impacting how robots grasp and transport these materials and how they interact in contact with human skin or other clothing layers.  %The frictional characteristics can be deduced through the angle of repose and the resulting trigonometric calculations, allowing for the derivation of the coefficient of friction.

These mechanical properties vary based on the production process the textile undergoes. They can be measured through standardized methods from the textile industry~\cite{cusick1968} or through procedures tailored to robotics applications~\cite{garcia-camacho2024standardization}. Their interdependent influence in cloth deformation poses fundamental challenges to the analysis and understanding of how these properties affect robotic manipulations.

% \al{unclear how the paragraph that follows is relevant in this part of the review.}
% Each of these mechanical properties is interdependent and varies based on the production process the textile undergoes. %These variations highlight the complexity of modeling textile objects for robotic manipulation, necessitating advanced algorithms that can adapt to the diverse mechanical behaviors exhibited by different textiles. %\dave{Can we modify this section to talk more about the papers that address these variations?}

%% Cloth sets radar chart figure:
% \begin{figure}
%   \centering
%   \includegraphics[width=\textwidth]{images/radar_chart_examples.png}
%         \caption{Representation of variation of textile properties for 3 cloth sets in literature. \al{Use a similar chart to introduce the factors of variations of textiles. NOTE: select 3 objects relevant for manipulation. Find their properties. and put them in the chart but not with the range but with the spectrum.}\TODO{Provisional figure. Decide what is the best way to represent the concept of variation (It would probably be better to not use the comparison chart of several cloth sets).
%         }}
%         \label{fig:radar_chart}
%         \vspace{-3pt}
% \end{figure}

%% Option 1: Examples of subjectives values of properties 

%% Option 2: Object examples radar chart figure
% \begin{figure}
%   \centering
%   \includegraphics[width=\textwidth]{images/objects_radar_chart2.png}
%         \caption{Representation of variation of textile properties for 4 cloth objects. Quantitative measures of the properties are obtained with the framework proposed in \cite{garcia-camacho2024standardization}. \TODO{Include in the figure the maximum values of each axis.}}
%         \label{fig:radar_chart}
%         \vspace{-3pt}
% \end{figure}

\subsection{Variations of Tasks}

% \al{In some way we would like to discuss to what extent these tasks are addressed.}

Manipulation of textiles spans a wide range of applications across domestic, healthcare, and industrial settings, including tasks such as laundry, tidying, dressing, sorting, and more. Each of these tasks presents unique challenges that necessitate the design and study of specific manipulation strategies that account for specific variations in the physical and mechanical properties of textiles discussed in the previous sections.

Following the taxonomy proposed by Mason~\cite{mason2001mechanics}, manipulation techniques can be characterized based on the nature of forces involved: kinematics, static forces, quasi-static forces, and acceleration forces. In the context of manipulating deformable objects, most current techniques can be broadly classified into two main categories~\citep{blanco2024towards}: quasi-static manipulation, which involves slow motions allowing static equilibrium, and dynamic manipulation, which involves motions that include acceleration forces. Physical properties like material, as well as shape and size, significantly affect both types of manipulations. On the other hand, while mechanical properties play a fundamental role in tasks requiring dynamic manipulations due to the influence of acceleration forces, tasks involving quasi-static manipulations are less dependent on mechanical properties.

For instance, tasks such as classification or sorting highly depend on physical properties such as the shape of the cloth, as well as the material the textile is composed of~\cite{sun2017single}. Similarly, tasks like flattening and folding rely heavily on understanding and manipulating the shape of the textile~\cite{sun2014heuristic,lee2021learning_one_hour,doumanoglou2016folding,hoque2022visuospatial}. In these cases, elasticity does not directly impact the outcome of the manipulation. However, stiffness and friction still play a role in the deformed state in which the object is observed. Regarding shape properties, flattening and folding tasks are typically performed on textiles laid flat on a surface, simplifying the manipulation process by reducing the need to consider more complex topographical features such as loops, holes, or cylindrical parts. In contrast, tasks such as assistive dressing~\cite{wang2023one} or hanging an apron on a hook~\cite{antonova2021dynamic} require a nuanced understanding and control of the textile's topological features, including connectivity, holes, voids, and spatial relationships~\cite{antonova2021sequential}. These tasks are inherently more complex due to the need to manipulate the textile in three-dimensional space and adapt to its changing configuration. Mechanical properties, such as elasticity and stiffness, play a crucial role in these scenarios, as they influence how the textile drapes, stretches and recovers from deformation.

\section{MODELING}
\label{sec:modeling}
%\todoinb{Hang - Complete discussion about unmodeled phenomenas}
%\todoinv{Yufei, David BM, Marco}
%\todoinor{David BM - Iterate over sim2real gap}
%\todoinor{Hang, Michael - Iterate over feedback}
%\todoinor{Hang, Michael - Shrink to 3 pages}

Accurate modeling physical properties of textile objects is crucial for robotic manipulation, computer graphics, and material sciences. In this section, we introduce modeling techniques commonly used in robotics, focusing on how they characterize various properties and their effectiveness in capturing them. Rather than elaborating on the technical aspects of these techniques, which are extensively covered in other reviews, we aim to provide an overview of their utility and the types of physical and mechanical variations they address. The reality gap within commonly used simulators is also covered with a discussion about currently unmodeled phenomena and techniques for real-world alignment. For an in-depth technical discussion on textile modeling techniques, we refer readers to~\cite{arriola2020modeling,yin2021modeling,hou2019review}.

\subsection{Physics-based Models}
\label{sec::model_physics}
%\TODO{Decide how to shrink this section}
%\dbm{Generally in each modeling technique there's no explanation of what type of variations can be addressed with each model, which was claimed in the intro before}
Physic-based models form the foundation for the most commonly used simulators in robotics. These models usually describe the cloth state through geometric representations such as particles or meshes. Physics-based simulation provides a reproducible and scalable mechanism to study physics-based cloth models and cloth dynamics across a range of environments and tasks. In what follows, we characterize different modeling techniques based on the representation used for the cloth. %In particular, we will discuss mesh-based, particle-based, and continuous domain models.

\textbf{Mesh-based} models, such as mass-spring-damper models, represent the cloth as a mesh of interconnected masses and springs. The physical interaction between masses intuitively reflects the variation of textile stiffness and elasticity. These models are a popular choice implemented in many simulators, for example, MuJoCo~\cite{todorov2012mujoco} and SOFA~\cite{faure2012sofa}. While these models are rather straightforward to implement, they are most suitable for simulating small deformation but not complex elastic effects with high fidelity. Additionally, tuning parameters for the springs and dampers to achieve realistic behavior can be relatively challenging, as these parameters do not directly correspond to physical meaning in the real cloth. 

\textbf{Particle-based} methods in cloth simulation represent the material as a collection of discrete particles, each defined by properties like position, velocity, and mass. These particles are interconnected through holonomic constraints that describe their interactions. A widely used approach within this framework is Position-Based Dynamics (PBD), as implemented in simulation engines such as FLEX~\cite{macklin2014unified,lin2021softgym}. PBD focuses on satisfying constraints related to stiffness, elasticity, and collision by directly adjusting the positions of particles, with subsequent velocity updates derived from these positional changes. The advantages of PBD include rapid simulation speeds, enhanced stability, and the ability to simulate inextensible textiles and shear deformation. The ability to easily incorporate a variety of constraints makes it possible to model plastic deformations, as well as interactions with fluids and rigid bodies under flexible frictional models~\cite{macklin2014unified}. While PBD offers visually plausible dynamics, it may not always align with physical realism. Moreover, the interpretation of simulation parameters into physical parameters, such as material modulus, remains challenging and often necessitates extensive tuning to achieve desired effects.
%\dbm{This end sentence seems to me more like a summary of PBD, but does not go to the point of cloth and variations (which we make in the intro of the section).}

% \dbm{Standard MPM combines a particle representation with a computational grid to solve the equations of motion. Do you want to include SoftMAC2024 here, or some previous work that has no application to robotics? (2021, "Revisiting Integration in the Material Point Method: A Scheme for Easier Separation and Less Dissipation" has simulation of cloth with MPM.)  } \TODO{MW: IS THIS TRUE? Complete discussion about MPM}\hy{I'm also a bit unsure about the statement here, and actually the distinction between mesh-based and particle-based as well. MPM is a hybrid in the sense of mixing Eulerian (grid) and Lagrangian (particles) viewpoints.} \yufei{If so, would it be better to distinct them based on the solver, e.g., mass-spring based, PBD, MPM, continuous domains? (I am not an expert in this so feel free to ignore if this is not clear or correct. )}
%By combining both representations, MPM is a very promising approach to model large and complex deformation observed in cloth.

\textbf{Continuous domain} models leverage the principles of continuum mechanics rather than relying on discrete representations of particles or meshes. This approach treats physical quantities of objects as continuous fields, providing a more physically precise depiction of material deformations. The underlying partial-differential equations are commonly solved with the finite element method, which divides the field domain into small elements linked as meshes. While this approach can be computationally intensive, model parameters have a clearer physical interpretation for textile properties such as strain, stress and their relation through Young's modulus.
%i.e., how the object deforms when forces are applied, how much deformation is received relative to the original configuration, and how much internal stress is applied to the material respectively. The ratio between stress and strain can be expressed in Young's modulus, while Poisson's ratio measures how much a material will expand in directions perpendicular to the direction of stretching. 
Robotics-relevant simulators relying on this modeling technique are SOFA~\cite{faure2012sofa}, Isaac Sim~\cite{macklin2019tog}, and Bullet~\cite{coumans2016pybullet} but specifically for deformable objects. Recently, the Material Point Method (MPM) \cite{mpm2016siggraph} emerges with a mix of particle representation, for which the particles carry material properties, while a spatial grid is used to compute forces and update the state of particles. By combining both representations, MPM is promising for rapid simulation of complex deformations observed in a cloth.
%\dbm{Add citation here to MPM, looks good.}
%\hy{I chose to move MPM under the umbrella of continuum because the original place seems to imply it is closer to mss/particle categories, while MPM is based on governing equation of continuum materials, see Jiang Chenfanfu's siggraph notes. Mesh-based and particle-based here are more "intuition-based" to me in that we cannot expect more accurate modelling by simply increasing the number/resolution of particles/meshes (PBD can be used under a continuum formulation as well though).}
% The recent SoftMAC2024 work~\cite{} deploys a hybrid approach between particle-based and mesh-based approaches. The employed Material Point Method (MPM) has the particles carry material properties, while the mesh connections is used to compute forces and update the state of particles

Besides forward prediction, \textbf{differentiable simulators} have recently gained popularity with efficient derivative evaluation through automatic-differentiation or adjoint methods~\cite{larionov2022estimating}. %~\cite{li2022diffcloth, larionov2022estimating}.
%\dbm{You're citing XPBD. Maybe mention that these simulators have been used to extend existing approaches}
The feature extends existing model-based approaches for efficient policy learning and parameter identification, finding applications in cloth-related tasks~\cite{qiao2020scalable,chen2023daxbench}.%~\cite{qiao2020scalable,chen2023daxbench,liu2023softmac,chen2024differentiable}. %Notable examples include DiffCloth~\cite{li2022diffcloth}, DiffSim~\cite{qiao2020scalable}, and XPBD~\cite{larionov2022estimating}, DAXBENCH~\cite{chen2022daxbench}.
%Among these, differentiable simulators utilizing adjoint methods~\cite{li2022diffcloth, larionov2022estimating, stuyck2023diffxpbd}, as opposed to conventional automatic differentiation \yufei{maybe add some citations here?}, may enable the use of larger integration steps and produce more stable gradients, thereby enhancing the efficacy of policy learning~\cite{Qiao2021Efficient, ainsworth2021fasterpolicy}.

% Some differentiable simulators: DiffCloth~\cite{li2022diffcloth}, DiffSim~\cite{qiao2020scalable}, DAXBENCH~\cite{chen2022daxbench}, XPBD~\cite{larionov2022estimating}, SoftMAC~\cite{liu2023softmac}
% DiPac~\cite{chen2024differentiable}.

% \TODO{Integrate Hang comment.}
% \hy{one thing might be worth to comment for robot practitioners is that differentiable sim relying adjoint methods \cite{li2022diffcloth, larionov2022estimating, stuyck2023diffxpbd} instead of common auto-diff may allow for larger integrating steps and better-behaved gradients for policy learning \cite{Qiao2021Efficient, ainsworth2021fasterpolicy}}

\subsection{Data-Driven Models}
\label{sec::model_data_driven}
Data-driven models have become a prominent alternative to traditional physics-based methods for simulating complex materials and deformations.  Compared to physics-based techniques, data-driven models offer greater flexibility in defining the state space for modeling cloth dynamics, as well as lower computational complexity and easier parallelization.

These models learn cloth dynamics directly from data, which can range from handcrafted observations, such as cloth key points, to raw images and 3D point clouds or a latent representation of these.  Image-based inputs have the benefit of not requiring knowledge of the 3D state of the cloth as the dynamics are learned either in the pixel space~\cite{hoque2022visuospatial} or in a latent representation of the image~\cite{lippi2020latent}. It also offers the advantage of working under partial observability~\cite{hoque2022visuospatial}. However, these often struggle with domain shifts that can occur with changes in camera positions, lighting, or background conditions, which impacts their generalization to new environments. Particle-based representations are often more robust to changes in visual conditions as they rely on 3D geometric representations such as particles or meshes rather than visual clues. These representations require specific architectures to efficiently capture local structures and handle data sparsity. Examples include PointNet++~\cite{qi2017pointnet++} for unordered point sets and Graph Neural Networks (GNNs)~\cite{scarselli2008graph} for mesh-based representations. 
%As for image-based models, the cloth dynamics can be either learned in the 3D state-space~\cite{} \yufei{why is image in the 3D state-space? Also should this sentence be moved earlier to the place where we discuss image-based inputs?} or in a lower-dimensional embedding~\cite{}.
%While\yufei{I thought the word 'while' is usually used to indicate a contrast which is not the case here} these methods\yufei{I guess you mean particle-based models? it is confusing since you talked about image-based models the sentence before} closely align with physics-based methods due to their state representation, 
Despite the advantage of particle-based representation in generalization, these models are generally more computationally intensive than their image-based counterparts. In addition, registering a mesh from partial depth data could also be non-trivial for textiles under large deformation.

The flexibility of data-driven models allows one to address variations of cloth properties by designing appropriate training schemes or conditioning the models on the object properties or a latent representation of these. Invariance to color changes is typically achieved during augmented training with domain randomization~\cite{wu2019learning}. In contrast, mechanical properties like stiffness and elasticity are commonly addressed through explicit or implicit conditioning. Explicit conditioning estimates these properties or their latent representations through perception and concatenates the estimation to the input model~\cite{longhini2023elastic}. Specifically for graph or mesh representations, properties like stiffness and elasticity can be induced as edge features in the mesh structure to bias model learning. Implicit conditioning, on the other hand, incorporates these properties by conditioning the model on a recent history of observations or employing techniques such as intuitive physics~\cite{longhini2024adafold}. These approaches allow models to adapt to changes in mechanical properties based on observed behaviors.

\subsection{Reality Gap in Cloth Models}
\label{sec::model_gap}
Different modeling techniques often need to account for a trade-off between computational cost and model accuracy. Trading accuracy for speed often leads to a gap between the dynamics in simulations and those in real-world scenarios. This gap between the model and how the physical cloth behaves in the real world is particularly relevant to real-world textile manipulation. 
% ~\yufei{This sentence can be cut to save space, not really relevant to textile}
% Current literature studied under-modelled effects for rigid object \cite{tiboni2023dropo} and the impact of Young's Modulus in sim-to-real soft robots. 
Here, we identify gaps between the assumptions made by commonly used robot simulation and real textile properties, as well as progress to align simulation to the real world. 
%The ``Reality Gap" refers to the differences that occur between the simulation and how the physical cloth behaves in the real world.
%As previously discussed, different modeling techniques have to strike a balance between the computational cost and accuracy of the models to be useful in a practical scenario. In the robotics community, computational efficiency is often favored over accuracy, adopting an approximation of the model or keeping specific phenomena fully unmodelled. 

\subsubsection{Unmodeled Phenomenas}
% Contributing factors to the realty gaps are often present in:
% \begin{enumerate}
%     \item Material Properties: The models rarely go down to the general building blocks of fabrics as laid out in section REF and do not accurately model things like fabric weave, thread count, or the response to real-world factors such as humidity, temperature, or wear and tear.
%     \item  Many real-life physical interaction phenomena that not only model cloth are notoriously difficult to model accurately such as air resistance, friction, and specifically for cloth the collision with other deformable or rigid objects.
%     \item The nonuniformity of cloth is a often completely ignored contributor to modeling cloth in simulators, full pieces of clothing consist often out of many different pieces sued together making it extremely challenging to capture all nuances introduced by this manufacturing process.
%     \item Partial tearing: Simulating the effects of ripping a cloth apart or adding holes to it is also extremely challenging to achieve computationally MABY ADD THE HANG CITATIONS HERE?
% \end{enumerate}
Robot simulators generally favour fast and visually correct behaviours at the expense of physical realism. Geometrically, textiles are often modelled as 2D objects without thickness. This requires extra care when the task concerns stacking or folding textiles of multiple layers. The mechanical modelling, be it physics-based or data-driven, rarely goes down to the basic building blocks of fabrics as laid out in Section~\ref{sec:background:textiles}. The simulation hence neglects aspects like fabric weave, thread count, as well as the response to real-world factors such as humidity, temperature, or wear and tear. In computer graphics, yarn-level simulation \cite{otaduy2014yarnlevel, sperl2021yarntog} has been studied to account for warp and weft interactions. However, this has not yet been integrated into robotics modeling pipelines. The simplification of simulating effects in smaller granularity also impacts the fidelity of how a cloth interacts with external objects. Textiles likely exhibit more complex sliding behaviours along different directions due to how they are constructed. This breaks the popular isotropic friction cone model in many simulators. 
%Yarn-level simulation has explored this to improve the contact realism \cite{pabst2009anisotropic, cirio2017yarnsliding}. 
Missing yarn-level simulation also influences modeling the tear-up of a cloth. Current robot simulators often assume invariant topology and uniformity about how a cloth is constructed. Graphics research has demonstrated the possibility of capturing these nuances with advanced simulation \cite{otaduy2014yarnlevel,meta2009tearing}. %The  limitation on modelling yarn details can also impact how the materials are sensed and hence image-based policies. Photometric characteristics of textile surface can be complex for the dependency on how are they weaved. Research from graphics  \cite{pabst2009anisotropic, textile2024CVPR} may be utilized to shrink the reality gap. 
Furthermore, another phenomena that is rarely included in cloth models is the aerodynamics of the textiles. Since most garments are very light, even in the absence of wind, the air that surrounds them has a critical impact on highly dynamic cloth motions \cite{coltraro2024validation}. Much research is still needed in order to have realistic, yet efficient physical models of cloth aerodynamics.

% \hy{Yarn-level \cite{otaduy2014yarnlevel, sperl2021yarntog} simulation has been studied in graphics but not really adopted in robotics simulators yet. so it would be hard to account for the properties due to warp and weft interactions. This may be also related to modeling the breaking behavior of textiles \cite{meta2009tearing, otaduy2014yarnlevel}. An-isotropic frictional contact that may be typical to cloth materials\cite{pabst2009anisotropic, cirio2017yarnsliding}. } 

%\dbm{I don't think this text fits into unmodeled phenomena. The visualization is not something unmodeled. }

% \hy{rendering might also be something robotics simulator falling behind the graphics frontier \cite{pabst2009anisotropic, textile2024CVPR}. this could be useful for shrinking the perceptual sim-to-real gap.}

%\subsubsection{Bridging the gap}
\subsubsection{Aligning Simulation to the Real World}
%%%%%%%%%%%%%%%%%%%%%%%%%%%%%%%%%%%%%%
%\al{refer to bruno siciliano review}
\label{sec::model_real2sim}
Physics simulation is essential for generating synthetic training data, exploring learned policies, and predicting the performance before real-world deployment. Thus, bridging the gap between the real world and simulation is crucial for generalizing manipulation skills to real-world textiles.  A common approach to reduce this gap involves tuning the simulation parameters to align with reality. This process is often referred to as real2sim, or system identification, and can be classified into three categories~\cite{arriola2020modeling}: gradient-based techniques, global optimization techniques, and neural network-based techniques. For gradient-based optimization, differentiable simulators backpropagate the error between the real-cloth state and the simulated one for system identification. These methods, however, often face challenges with discontinuous loss landscapes, particularly in scenarios involving deformable objects. For scenarios where the analytical model is not differentiable, global optimization techniques such as Bayesian Optimization (BO) and Covariance Matrix Adaptation Evolution Strategies (CMA-ES) are often used without the need for gradients~\cite{blanco2024benchmarking, erickson2018deep}. Global search methods can be computationally intensive and may struggle with scalability in high-dimensional spaces. A combination of the two has been proposed in~\cite{antonova2023rethinking}, where they combine global search using BO with a semi-local search to retain the benefit of gradient-based optimization but integrate BO for the parts of the landscape that are intractable for gradient descent alone. Another approach is to combine Bayesian inference with neural networks to infer simulation parameters~\cite{ramos2019bayessim}. One of the major benefits of this class of methods is capturing uncertainty in parameter estimates.

\section{PROPERTIES PERCEPTION}
\label{sec:perception}

The perceptual capabilities of robots encompass a variety of skills, including state estimation, segmentation, tracking, recognition, classification, and the identification of appropriate grasping points on cloth items. A comprehensive review of perception for grasping is presented in \cite{jimenez2020perception}. Additionally, discussions on state estimation, parameter identification, and detection are provided in \cite{yin2021modeling}, where the focus was not only on textiles but on deformable objects in general. This section will specifically delve into the identification of textile properties, aiming to focus on perceptual capabilities that enable the generalization and adaptability of robots to variations in the textile properties introduced in prior sections.

% Although robots in human environments will almost always be working with uncertainty due to the always-changing world, perceptual systems have the potential to reduce this uncertainty and enable robust autonomous operation by providing sensory feedback to the robot~\cite{kemp2007challenges}. This feedback might be obtained from audio, visual, tactile, and force-sensing modalities, each providing different information that might be relevant to the underlying task. While audio, tactile, and force measurements allow the identification of physical properties such as surface friction~\cite{gao2023controllable}, elasticity, or construction techniques~\cite{luo2018vitac,longhini2023elastic}, vision provides global information about the shape of the object. As such, the choice of the sensing modality is highly dependent on the downstream task. %Moreover, methods can leverage a combination of multiple modalities. 

Although robots in human environments constantly face uncertainty about object properties due to changing conditions, perceptual systems can reduce this uncertainty by providing sensory feedback~\cite{kemp2007challenges}.
Perception can be passive, requiring no physical interaction with the objects of interest. Properties such as color, shape, and material can be estimated using passive perception~\cite{isola2015discovering}. 
However, properties such as elasticity and friction cannot be observed in static scenarios. Therefore, interactive perception is necessary to obtain accurate estimates of these properties ~\cite{bohg2017interactive}. %The benefit of interactive perception is its ability to disambiguate between properties that are not observable from static scenarios, enabling the robot to reduce uncertainty. Different actions will result in different effects, and hence, robots can learn about their environment faster by selecting more informative actions.  
%For instance, actions such as pulling can provide valuable information about a material's elasticity, while sliding a haptic sensor over a surface is more informative about friction properties, allowing the robot to tailor its actions to specific property estimations. Thus,  appropriately coupling sensing and exploratory actions is of fundamental importance.

\begin{figure}
  \centering
  \includegraphics[width=1.\textwidth]{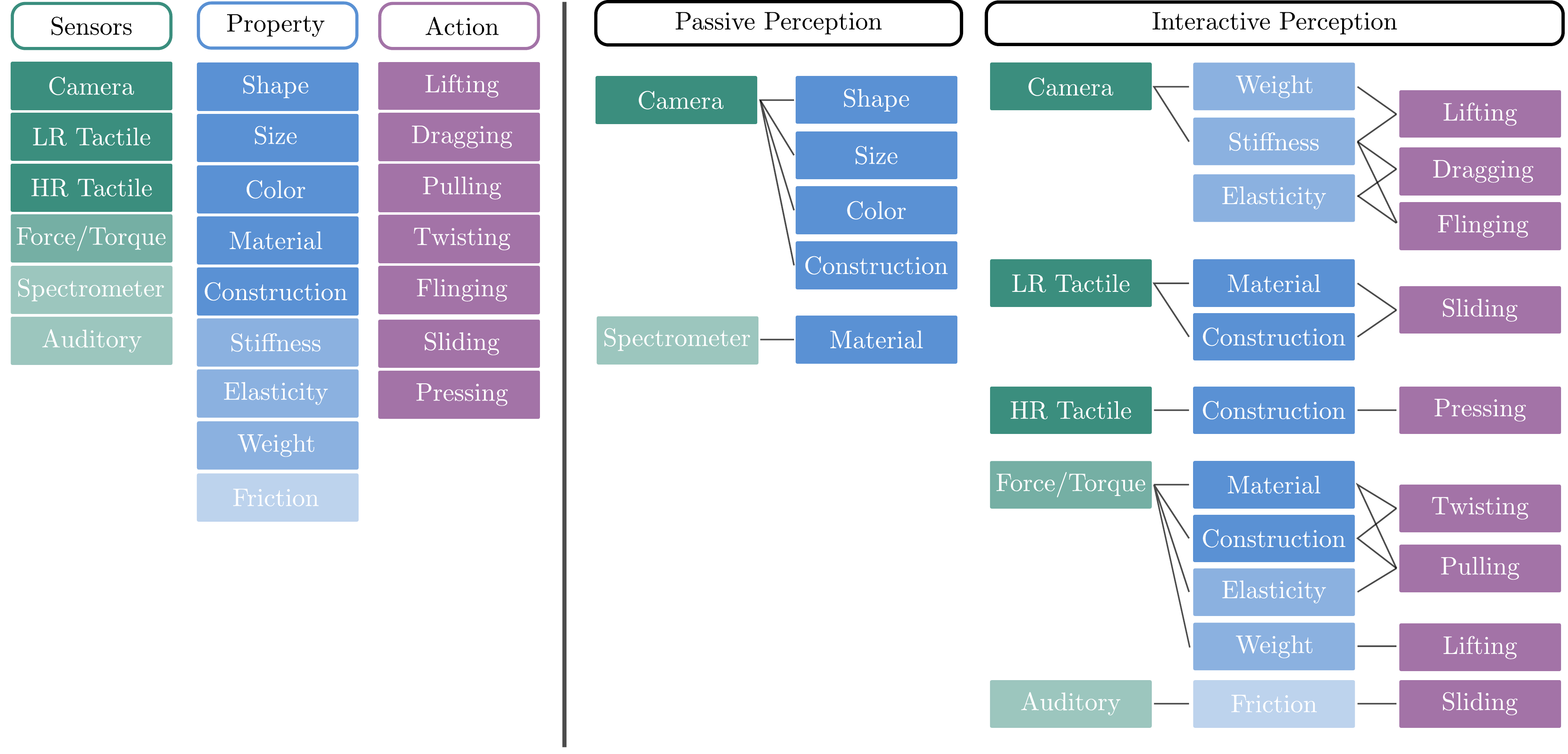}
        \caption{\textbf{Overview of sensors, properties, and actions in passive and interactive perception for textile manipulation}. On the left, the figure presents a list of sensors, properties, and actions. LR and HR for tactile sensors stand for low resolution and high resolution, respectively. The variation in opacity of the boxes reflects to what extent the sensor or the property has been explored for textile perception. On the right, the figure shows connections between sensors, properties, and actions based on the current literature. This side of the figure showcases how passive perception methods, using sensors like cameras and spectrometers, estimate properties that do not require physical interaction. In contrast, interactive perception integrates actions to improve the robot's understanding of mechanical properties, leveraging specific actions coupled with sensor feedback to accurately perceive properties such as weight, stiffness, and elasticity. }
        \label{fig:perception_table}
        \vspace{-3pt}
\end{figure}

%\al{Missing some thoughts about the actions needed for exploration and interaction.}
Designing the perception process relies on several key factors: the relevant properties for the tasks at hand, the available sensors, and the manipulator configuration. Figure \ref{fig:perception_table} provides an overview of various design choices for estimating textile properties. This table also represents the literature discussed in the remainder of the section.

% Designing the perception process along the interaction depends on the relevant properties for the tasks at hand, available sensors, and manipulator settings. A current overview of different design choices for textile properties estimation is showcased in Fig.~\ref{fig:radar_chart}. \al{Improve the reference to the table.} %In the remaining section, we will review the literature on the perception of properties for textiles. 

\subsection{Physical Properties Perception}
As noted in Section~\ref{sec:physical_properties}, physical properties like size, shape, weight, color, fabric material, and construction technique can often be observed and measured without external manipulation. Particularly, size and shape can be effectively inferred by observing textiles spread over a flat surface and utilizing contour or segmentation techniques~\cite{wang2011perception}. Data-driven techniques such as landmark detection or end-to-end classifiers have also been explored for shape classification~\cite{shajini2021improved,liu2016deepfashion}. However, textiles frequently arrive in deformed or crumpled states, complicating direct visual assessment. While some approaches have explored feature extraction processes to classify clothes from highly crumpled configurations~\cite{sun2017single}, interactive perception is often necessary to enhance perceptual accuracy~\cite{duan2022continuous}. %,martinez2019continuous}. 
Lifting interactions, for example, are commonly employed to infer the shape of the textile better and facilitate its classification by garment type ~\cite{willimon2011classification,corona2018active}. %,kita2009clothes}. %Alternatively, involving flattening manipulation strategies helps reach a configuration where there is no ambiguity about the shape and the cloth size\cite{}.

The estimation of cloth weight has been relatively underexplored within the robotics community. Pioneering investigations, primarily from the computer graphics domain, have demonstrated the potential of estimating the weight per area of textiles through video analysis of the object being influenced by external forces like wind~\cite{bouman2013estimating,runia2020cloth}. This interaction could be analogously implemented in robotic systems as a flinging action~\cite{ha2022flingbot}, although this specific application has yet to be explored for weight estimation in a robotic context. Current approaches in the robotic community estimate cloths weights through the estimation of external forces/torques felt by the robot during motion tasks such as lifting~\cite{colome2013external}. An alternative approach is to rely on a similarity network to infer cloth weight from lifting interactions using ground-truth cloth weight as supervision~\cite{duan2022continuous}. % was proposed in~\cite{duan2022continuous}, where the method relies on a similarity network to infer cloth weight from lifting interactions using ground-truth cloth weight as supervision. %Additionally, employing force-torque sensors under gravitational influence presents a straightforward approach to be investigated. 

Textile material and construction techniques have received considerable attention in the robotic community to explore combinations of visual and haptic sensing specifically~\cite{strese2019haptic}. %$Microscopic cameras provide a detailed view of the textile, revealing yarn features specific to each material. While integrating microscopes might be challenging, 
Photometric stereo sensors can reconstruct the surface of a region of the cloth at the yarn level and exploit the 3D pattern to infer the material and construction properties of the object~\cite{kampouris2016multi}. Alternatively, high-resolution tactile sensing such as GelSight~\cite{yuan2017gelsight} has shown to be relevant for both material classification and construction technique classification of fabrics. These sensors provide high-resolution images,  by pressing the sensor on the cloth~\cite{yuan2018active}. More specialized robot sensors, such as micro spectrometers, offer detailed insights into the yarn material composing the fabric \cite{erickson2020multimodal}.

Since both material and construction techniques influence the mechanical properties of cloth, interactive perception can provide additional insights to identify these properties accurately. %As a result, researchers have explored using interactive perception with various sensors to enhance material and construction technique estimation.  L
Low-resolution tactile sensors like the BioTac, coupled with a contact microphone, have been explored to infer the material of textiles by exploring the feature of the signal recorded from a sliding interaction over the textile~\cite{strese2019haptic}. Additionally, force-torque sensors used in conjunction with pulling and twisting actions have provided further insights into materials and construction techniques by observing how the textile responds to different mechanical stresses~\cite{longhini2021textile}.  %These interactive techniques help to refine the estimation of material and construction by relating them to the influenced mechanical properties.

\subsection{Mechanical Properties Perception}

Mechanical properties of textiles, such as stiffness and elasticity, are critical in determining how a fabric will behave under various types of stress and strain. These properties are traditionally measured using sophisticated and costly systems that are standard in the textile industry. A prominent example is the Kawabata Evaluation System for Fabrics, which quantifies a fabric's response to controlled forces by measuring parameters such as stress-strain behavior at maximum load for the specific material being tested~\cite{kawabata1989fabric}. Additionally, the computer graphics community has conducted in-depth studies of cloth elasticity using setups that combine pulling interactions with force and visual observations \cite{wang2011data,miguel2012data}. However, these setups are typically unsuitable for robotic applications due to their lack of real-time interaction and computational capabilities.

In the robotics community, stiffness and elasticity have been explored using a combination of force-torque sensors and camera observations during robotic pulling interactions~\cite{longhini2023elastic}.
%Stiffness and elasticity directly relate to the deformation properties of the textile under stress. To explore these characteristics, researchers have utilized a combination of force-torque sensors and camera observations during robotic pulling interactions \cite{longhini2023elastic}. 
While this approach directly measures how textiles respond to controlled mechanical stresses, significant research focuses on methods using only camera observations to detect textile deformation under external forces like gravity or wind. Specifically, work using cameras can be broadly divided into two categories:  1) optimizing simulation parameters to reflect the behavior of real-world objects accurately and 2) applying external forces and observing resultant shape changes. The first category leverages 3D observation, such as point clouds, to optimize simulation parameters such as elasticity, bending, and stiffness~\cite{zheng2024differentiable,sundaresan2022diffcloud}. The second category instead includes methods from the computer vision community, where 2D images are used to assess how a textile drapes under gravity or moves under external forces. These observations focus on visual features such as wrinkles or the dynamic motion of textiles over time~\cite{larionov2022estimating,duan2021learning,yang2017learning}. Notably, the studies of this last category do not involve robots, indicating a potential new research direction to replicate these techniques in robotic systems.
% Specifically, work using cameras can be broadly divided into two categories:  1) applying external forces and observing resultant shape changes, and 2) optimizing simulation parameters to reflect the behavior of real-world objects accurately. The first category includes methods where cameras are used to assess how a textile drapes under gravity or how it moves under external forces such as wind. These observations focus on visual features from 2D images, such as wrinkles or the dynamic motion of textiles over time\cite{larionov2022estimating,duan2021learning,yang2017learning}. \al{most of these studies do not involve robots, but in some ways, this could be replicated with robots. It is an interesting research direction.}. Instead, the second class of methods leverages 3D observation, such as point clouds, to optimize simulation parameters and rely on these parameters as an estimation\cite{zheng2024differentiable,sundaresan2022diffcloud}. In this line of research, actions such as lifting, dragging, and flinging have been explored to exploit external forces such as gravity or surface friction and observe the resulting textile deformation.  
% In contrast, friction influences deformation only when the textile is in contact with another surface, such as a table, other cloth, or human skin. Robotic techniques for friction estimation have involved sliding low-resolution haptic and auditory sensors over cloth to extract relevant features about friction coefficients\cite{chu2013using}. 
Friction, on the other hand, remains largely underexplored. While some work has investigated haptic adjectives related to friction, direct estimation techniques are limited, with efforts mainly focusing on using low-resolution haptic and auditory sensors to infer friction-related features~\cite{chu2013using}.

\section{MANIPULATION}
\label{sec:manipulation}

Robotic manipulation in human environments often confronts highly unstructured settings, necessitating the ability to manipulate objects with varied characteristics. In Section~\ref{sec:background}, we reviewed textile variations in terms of properties and tasks, and highlighted the complexity of addressing the variations of these real-world textiles. These complex environments underscore the importance of endowing robots with robust adaptation capabilities. Here, adaptation refers to the ability of the robot to adjust its strategies in response to changes in its operating environment or variations in the objects it manipulates.
Effective manipulation skills require adaptability to both previously observed and novel variations of textile properties~\cite{kemp2007challenges}.

Successfully handling different clothing variations can be achieved through two primary mechanisms: generalization and adaptation~\cite{cui2021toward}. Generalization assumes that the distribution of knowledge acquired in the source domain will encompass the distribution of the target domain, resulting in a successful transfer. This assumption allows the robot to handle known variations, but it may not hold in the presence of novel variations typical of diverse human environments, where previously learned knowledge may fall short. A common framework to achieve generalization is domain randomization~\cite{wu2019learning}. In contrast, adaptation involves dynamically modifying the manipulation strategy or dynamics model by taking into account environmental changes.
This adaptation can be achieved by parameterizing the dynamics model~\cite{longhini2024adafold,longhini2023edonet} or the manipulation policy with the estimated cloth properties. 

In the following sections, we group common manipulation strategies into three categories, provide a high-level overview of each, and discuss in more detail their adaptability towards different textile properties and task variations.
A general overview of these methods is given in Table~\ref{tab:textile_methods}, including the tasks these are applied to as well as the variations of cloth physical and mechanical properties that are evaluated.

% \begin{sidewaystable}[h]
\begin{table}[b]
\centering
\caption{
Overview of textile variations handled by methods in the literature. Methods that do not assess their generalization or adaptation over textile variations are marked with \rxmark,
% while  \gcmark and (\gcmark) are used to characterize 
methods that address textile variations explicitly with \gcmark, and with (\gcmark) when not explicitly mentioned but the evaluated textiles contains such variations. 
% We consider evaluations conducted in the real world with at least one variation of the physical or mechanical properties of textiles.
}
\label{tab:textile_methods}
\tabcolsep7.5pt
\begin{tabular}{@{}l|c|c|ccccc|cccc@{}}
\hline
Ref. & \# O & Task & \multicolumn{5}{c|}{Physical Properties} & \multicolumn{4}{c}{Mechanical Properties}  \\
\cline{4-12}
& & & SH & SZ & CLR & M & CT & W & E & ST & F \\
\hline
\rowcolor[gray]{0.9} \multicolumn{12}{c}{Model-Based} \\
\hline
\cite{lippi2020latent} & 1 & Fo & \rxmark& \rxmark & \rxmark & \rxmark & \rxmark &\rxmark & \rxmark & \rxmark & \rxmark  \\
\cite{ma2022learning} & 1 & Fo/Fl & \rxmark& \rxmark & \rxmark & \rxmark & \rxmark & \rxmark & \rxmark & \rxmark & \rxmark \\
\cite{longhini2024adafold} & 6 & Fo & \rxmark & \gcmark & \gcmark & \gcmark & \gcmark & \gcmark & \gcmark & \gcmark & \rxmark \\
\cite{li2015folding} & 7 & Fo & \gcmark& \gcmark & \gcmark & \rxmark & (\gcmark) & (\gcmark) & \gcmark & \gcmark & \gcmark \\
\hline
% Flattening
\cite{hoque2022visuospatial} & 1 & Fl & \rxmark& \rxmark & \rxmark & \rxmark & \rxmark & \rxmark & \rxmark & \rxmark & \rxmark \\
\cite{yan2021learning} & 3 & Fl & \rxmark& \rxmark & \gcmark & \rxmark & \rxmark & \rxmark & \rxmark & \rxmark & \rxmark \\
\cite{lin2022learning} & 3 & Fl & \gcmark & \gcmark & \gcmark & (\gcmark) & \rxmark & (\gcmark) & (\gcmark) & (\gcmark) & (\gcmark) \\
\cite{huang2022mesh} & 5 & Fl & \gcmark & \gcmark & \gcmark & (\gcmark) & \rxmark & (\gcmark) & (\gcmark) & (\gcmark) & (\gcmark) \\
% Dressing
\hline
\cite{erickson2018deep} & 1 & D & \rxmark & \rxmark & \rxmark & \rxmark & \rxmark & \rxmark & \rxmark & \rxmark & \rxmark \\
\hline
\rowcolor[gray]{0.9} \multicolumn{12}{c}{Model-Free} \\
\hline
\cite{matas2018sim, salhotra2022learning, lee2021learning_one_hour} & 1 & Fo & \rxmark& \rxmark & \rxmark & \rxmark & \rxmark & \rxmark & \rxmark & \rxmark & \rxmark \\
\cite{tsurumine2019deep} & 2 & Fo & \gcmark& \gcmark & \gcmark & \rxmark & \rxmark & \rxmark & \rxmark & \rxmark & \rxmark \\
\cite{weng2022fabricflownet} & 3 & Fo & \gcmark& \gcmark & \gcmark & \rxmark & \rxmark & \rxmark & \rxmark & \rxmark & \rxmark \\
\cite{hietala2022learning} & 3 & Fo & \rxmark& \rxmark & \gcmark & \gcmark & (\gcmark) & \gcmark & (\gcmark) & (\gcmark) & \rxmark \\
\cite{avigal2022speedfolding} & 3 & Fo & \gcmark & \rxmark & \gcmark & \rxmark & \rxmark & \rxmark & \rxmark & \gcmark & \rxmark \\
\cite{lee2022learning} & 5 & Fo & \rxmark & \rxmark & \gcmark & \rxmark & \rxmark & (\gcmark) & \rxmark & \rxmark & \rxmark \\ 
\cite{petrik2019feedback} & 10 & Fo & \rxmark & \rxmark & \gcmark & \gcmark & \rxmark & \rxmark & (\gcmark) & (\gcmark) & (\gcmark) \\
\cite{ganapathi2021learning} & 10 & Fo/Fl & \gcmark & \gcmark & \gcmark & \rxmark & \rxmark & (\gcmark) & \rxmark & \rxmark & \rxmark \\
\cite{xue2023unifolding} & 20 & Fo & \gcmark & \gcmark & \gcmark & \gcmark & (\gcmark) & (\gcmark) & (\gcmark) & (\gcmark) & (\gcmark) \\
\hline
% Flattening
\cite{seita2020deep} & 1 & Fl & \rxmark& \rxmark & \rxmark & \rxmark & \rxmark & \rxmark & \rxmark & \rxmark & \rxmark \\
\cite{wu2019learning} & 3 & Fl & \rxmark& \rxmark & \gcmark & \rxmark & \rxmark & \rxmark & \rxmark & \rxmark & \rxmark \\
\cite{ha2022flingbot} & 3 & Fl & \gcmark& \gcmark & \gcmark & \rxmark & \rxmark & \rxmark & \rxmark & \rxmark & \rxmark \\
\cite{blancomulero_2023_qdp_unfolding} & 3 & Fl & \gcmark& \gcmark & \gcmark & \gcmark & \rxmark & \gcmark & \gcmark & \rxmark & \rxmark \\
\hline
% Dressing
\cite{tamei2011reinforcement,zhang2022learning} & 1 & D & \rxmark & \rxmark & \rxmark & \rxmark& \rxmark & \rxmark & \rxmark & \rxmark & \rxmark \\
\cite{sun2024force} & 2 & D & \gcmark& \gcmark & \gcmark & (\gcmark) & \rxmark & (\gcmark) & (\gcmark) & (\gcmark) & (\gcmark) \\
\cite{joshi2019framework} & 3 & D & \rxmark & \rxmark & \gcmark & \rxmark& \rxmark & \rxmark & \rxmark & \rxmark & \rxmark \\
\cite{wang2023one} & 5 & D & \gcmark& \gcmark & \gcmark & (\gcmark) & \rxmark & (\gcmark) & (\gcmark) & (\gcmark) & (\gcmark) \\
% Bed Making
\hline
\cite{seita2019deep}  & 3 & B & \rxmark & \rxmark & \gcmark & (\gcmark) & \rxmark & \rxmark & \rxmark & \rxmark & \rxmark \\
\hline

\rowcolor[gray]{0.9} \multicolumn{12}{c}{Heuristic-Based} \\
\hline
\cite{miller2012geometric}  & 4 & Fo & \gcmark & \gcmark & \gcmark & (\gcmark) & \rxmark & \rxmark & \rxmark & \rxmark & \rxmark  \\
% \hline
\cite{doumanoglou2016folding}  & 12 & Fo & \gcmark & \gcmark & \gcmark & (\gcmark) & \rxmark & (\gcmark) & (\gcmark) & (\gcmark) & (\gcmark) \\
% \hline
\cite{maitin2010cloth} & 25 & Fo & \rxmark & \gcmark & \gcmark & \gcmark & \rxmark & (\gcmark) & (\gcmark) & (\gcmark) & (\gcmark) \\ 
\hline 
\end{tabular}
\flushleft
\#O:Number of test objects, Fo:Folding, Fl:Flattening, D:Dressing, B:Bed Making, SH:Shape, SZ:Size, CLR:Color, M:Material, CT:Construction Technique, W:Weight, E:Elasiticy, ST:Stiffness, F:Friction. 
%\yufei{for~\cite{maitin2010cloth}, no explicit mention of the mechanical properties, but there are 25 towels tested, so I assume they will be different? Same for~\cite{miller2012geometric}, it used a towel, shorts, short and long-sleeved tshirt. I assume the mechanical properties will be different across these 4 categories?}
% \dbm{In ~\cite{blancomulero_2023_qdp_unfolding} we evaluated 3 different cloths with different shapes, colors, sizes and also mechanical properties.}
\end{table}
% \end{sidewaystable}

\subsection{Model-based Manipulation}
%intro
Model-based manipulation approaches refer to the class of methods that use a model of the textile to generate the manipulation strategy. The manipulation actions are usually obtained by planning with the model~\cite{erickson2018deep,lin2022learning,huang2022mesh}. 
These approaches can be categorized based on the methods used to construct the model, namely into analytical models and data-driven models.

\subsubsection{Analytic Models}
% \TODO{Shrink, reuse section 3.1 as much as possible. }

Analytic models, or physics-based models, build the textile model based on the laws of physics.
Please refer to Section~\ref{sec::model_physics} for a more detailed discussion on the popular analytic models and the textile properties they can model.
When using these models for manipulating a textile object, the initial step typically involves aligning the model dynamics to the real world to match the dynamics of the target textile~\cite{li2015folding, luque2024mpc_dynamic_cloth}. %; please refer to Section~\ref{sec::model_real2sim} for common real-to-sim techniques.
% tuning the model parameters to ensure that the model dynamics match those of the target textile~\cite{li2015folding, luque2024mpc_dynamic_cloth}. This process is known as system identification, or real-to-sim. Please refer to Section~\ref{sec::model_real2sim} for common real-to-sim techniques.
Analytic models have the advantage of being applicable to different tasks as the model is usually task-agnostic, as long as the textile dynamics required by the task can be accurately modelled by the chosen analytic model.

There are two major drawbacks of using analytic models for manipulation. First, their generalization is limited by the accuracy of the physics model, which, as discussed in Section~\ref{sec:modeling}, is often only an approximation of real cloth dynamics~\cite{blanco2024benchmarking}. 
Second, each new object requires a separate system identification process, which can be tedious and difficult to scale for adapting to a great variability of textile mechanical properties.
% \dbm{Do you know of any analytical model method that does adaptation??}

\subsubsection{Data-driven Models}
% \yufei{I feel section 3.2, especially the second paragraph, is already well-written and covers most of the things we might want to discuss about how data drive methods deal with variations. I think we can just refer to that paragraph for saving space instead of trying to re-iterate them here.}
% \dbm{I agree that we can refer to most things that are in section 3.2, but some of the references that are here are not covered in that section, nor the effect in manipulation, which is covered here.}
Data-driven models learn from data about interactions with textiles. There can be many state representations for data-driven models as laid out in Section~\ref{sec::model_data_driven}.
The choice of the state representation and model architecture can greatly affect the generalization ability of the model towards different properties, such as the shape and texture of the textile. A particle-based state representation combined with graph neural networks~\cite{lin2022learning, huang2022mesh} has been shown to generalize better towards different shapes, geometries, and mechanical properties of the cloth compared to images~\cite{hoque2022visuospatial} or latent-based state representations~\cite{lippi2020latent, yan2021learning}, as the particle-based state representations align better with the underlying cloth physics. Using depth images~\cite{ma2022learning} or point clouds~\cite{lin2022learning, huang2022mesh, puthuveetil2023robust} as the state representation also naturally makes the model invariant to the color and visual texture of the textiles. Due to the large amount of data needed for learning the model, most works use a simulator to generate the interaction data for learning the model~\cite{ma2022learning, hoque2022visuospatial, longhini2024adafold,yan2021learning,lin2022learning, huang2022mesh}, with a few that does so in the real world~\cite{lippi2020latent}.

As discussed in Section~\ref{sec::model_data_driven}, data-driven models often address variations of cloth physical properties
% , such as shape, color and texture, 
through domain randomization~\cite{ hoque2022visuospatial, yan2021learning, lin2022learning,huang2022mesh}.
% which randomizes these properties when training the model, so the trained model can be robust to such variations.
However, as current simulators~\cite{todorov2012mujoco, macklin2014unified, coumans2016pybullet} struggle in modeling a wide range of mechanical properties of the textiles, the generalization towards diverse mechanical properties that can be achieved through domain randomization remains limited. 
Still, recent work has shown that integrating a model conditioned on a latent representation of mechanical properties within a feedback-loop framework enables a model learned in simulation to adapt to textiles with diverse mechanical properties in the real world~\cite{longhini2024adafold}.  
% Still, one recent work has demonstrated that by conditioning the model on a latent representation
% % (encoded from a history of observations) 
% of the mechanical properties of the textile, models learned with only a few variations of mechanical properties in simulation can adapt to textiles with diverse mechanical properties in the real world~\cite{longhini2024adafold}. 

Most works in the literature learn an individual model for each type of manipulation task, e.g., assistive dressing~\cite{erickson2018deep}, cloth folding~\cite{ ma2022learning, longhini2024adafold}, cloth smoothing~\cite{ma2022learning, yan2021learning, huang2022mesh} or blanket covering~\cite{puthuveetil2023robust}. A few works have demonstrated that the learned model can be used to perform two tasks such as cloth folding and smoothing~\cite{ ma2022learning, hoque2022visuospatial, lin2022learning}. Learning a single model that can generalize to multiple manipulation tasks remains underexplored.

\subsection{Model-free Manipulation}

Model-free manipulation approaches directly map the state or sensory observation of the textile to the manipulation action without having a model in the loop during inference. 
Compared to model-based approaches, they require no prior knowledge of the textile as no model needs to be constructed.
There are two main approaches: reinforcement learning and imitation learning.     

\subsubsection{Reinforcement Learning}
\label{sec:manip-rl}

To apply reinforcement learning (RL) to a textile manipulation problem, the manipulation problem needs to be formulated as a Markov Decision Process (MDP) or a Partially Observable MDP (POMDP). The key in the formulation is the design of the state, action, and reward space of the MDP. 
As in the case of model-based methods, common choices for state representations of textiles include manually defined features such as key points on the cloth~\cite{petrik2019feedback, tamei2011reinforcement}, or results from a perception system such as an image~\cite{matas2018sim, wu2019learning,  hietala2022learning} or point cloud~\cite{sun2024force, wang2023one} of the textile. 
Common actions for manipulating textiles include action primitives such as pick-and-placing~\cite{wu2019learning}, dragging~\cite{avigal2022speedfolding}, flinging~\cite{ha2022flingbot}, or raw actions that control the delta movement or velocity of the robot end-effector~\cite{matas2018sim, wang2023one, hietala2022learning}.  
The reward function defines the desired outcome of the manipulation task based on the state of the environment. 

Given the formulated MDP, reinforcement learning algorithms can be used to find a policy that maximizes the expected accumulated rewards, thus learning the necessary manipulation skills.
Most works formulate the textile manipulation problem as a multi-step MDP~\cite{matas2018sim, wang2023one,wu2019learning,  hietala2022learning, blancomulero_2023_qdp_unfolding} and learn a policy to maximize the accumulated reward, while some works~\cite{ha2022flingbot, avigal2022speedfolding} formulate the problem into a bandit problem (an MDP with only 1 step), learn the reward function (usually in the form of a spatial-action value map~\cite{lee2021learning_one_hour, ha2022flingbot, avigal2022speedfolding}), and choose the best action as the one that maximizes the learned 1-step reward. 
Due to the large amount of data required by an RL algorithm, the need for a reward function, and the need to periodically reset the environment, such methods usually train a policy in a simulator and perform sim2real transfer~\cite{matas2018sim, wang2023one, wu2019learning,  hietala2022learning, petrik2019feedback, blancomulero_2023_qdp_unfolding} with optional real-world fine-tuning~\cite{ha2022flingbot}.
A few works~\cite{lee2021learning_one_hour, tsurumine2019deep, avigal2022speedfolding} directly train in the real world, where the reward can be automatically computed from real-world perception systems~\cite{avigal2022speedfolding}, and the environment can be  automatically~\cite{lee2021learning_one_hour, avigal2022speedfolding}, or manually~\cite{tsurumine2019deep} reset.  

Domain randomization is still the key technique for model-free RL methods to achieve generalization towards diverse textile mechanical and physical properties.
In domain randomization, properties of the textile and environment are randomized during the training process, so the resultant policy generalizes to all varied properties.
The randomized quantities can include the textile's shape~\cite{wang2023one, ha2022flingbot}, size~\cite{matas2018sim, wang2023one}, location and orientation~\cite{matas2018sim, wang2023one}, texture and lighting~\cite{matas2018sim, wu2019learning, hietala2022learning}, and mechanical properties~\cite{wu2019learning,hietala2022learning,  petrik2019feedback}. 
To achieve more informed domain randomization, contrastive learning can be used to compare pairs of real and simulated garment observations to learn a similarity metric~\cite{zhang2022learning}, which is used to tune the simulation parameters to align the simulation and real-world garment observations. 
Again, one caveat of domain randomization is that the range of mechanical properties that can be randomized is limited by the fidelity of the simulator. %,  which becomes a significant limitation as most model-free RL methods train the textile manipulation policy within the simulator. I
%f the target object to manipulate in the real world has a mechanical property that exceeds the range the simulator can simulate, the simulation-trained policy could still fail on this object even when training with randomized properties in simulation. 
There has been little work in the literature that explores addressing variations of textile properties via adaptation.

% then we talk about tasks where model-free RL has been applied.
In terms of tasks, model-free RL methods have been applied to many textile manipulation problems including cloth smoothing and flattening~\cite{lin2021softgym, wu2019learning, ha2022flingbot, avigal2022speedfolding, blancomulero_2023_qdp_unfolding}, folding~\cite{matas2018sim, lee2021learning_one_hour, tsurumine2019deep, hietala2022learning, avigal2022speedfolding}, placing~\cite{lin2021softgym}, hanging~\cite{matas2018sim, antonova2021dynamic}, blanket covering~\cite{puthuveetil2022bodies}, and assistive dressing~\cite{sun2024force, wang2023one,  tamei2011reinforcement, zhang2022learning}. 
All these works learn an individual policy for each task; the goal of %, and it remains highly unexplored in terms of 
learning a single policy over multiple tasks or that can generalize to different manipulation tasks remains highly unexplored. 

\subsubsection{Imitation Learning}

To apply imitation learning (IL) to textile manipulation problems, a dataset of expert demonstrations that solve the manipulation task needs to be first collected. %The expert demonstrations are usually in the form of paired states and expert actions taken in the states. 
Such demonstrations can be collected by a human~\cite{lee2022learning, xue2023unifolding, joshi2019framework} or using a scripted policy~\cite{seita2021learning, weng2022fabricflownet, seita2020deep, jia2019cloth}. 
%As in RL, IL approaches need to formulate the proper state and action representations for the manipulation problem. 
The most standard IL algorithm is behavioral cloning %, which trains the manipulation policy via supervised learning to predict the expert actions given state inputs
~\cite{lee2022learning, xue2023unifolding, seita2020deep}. Another approach is to directly map the actions in the demonstrations to the test object via learned correspondences~\cite{ganapathi2021learning}.

The generalization ability of IL approaches highly depends on the amount of variations presented in the demonstrations. 
An extensive dataset that covers diverse objects with varied properties is essential for learning a policy that can generalize across different textiles. 
Many works that use IL for cloth manipulation show limited generalization towards different textile properties, demonstrating their method on a fixed textile~\cite{seita2021learning, salhotra2022learning, seita2020deep, jia2019cloth}. 
Some works use depth images as observations, so the policy can be invariant to visual features such as color and texture%, and is shown to achieve generalization towards 2-4 new textiles with different textures
~\cite{weng2022fabricflownet, lee2022learning, seita2019deep}. 
When the collected demonstrations are diverse enough, a stronger generalization can be achieved: Xue et al.~\cite{xue2023unifolding} collected a dataset with hundreds of simulation shirts and 40 real-world shirts, and showed that the imitation policy can generalize to 20 real-world shirts with diverse shapes and materials.  One caveat of IL methods is that collecting a sufficiently diverse and extensive set of demonstrations can be human labor-intensive. 
%Similarly, very few works have explored combining imitation learning with adaptation-based methods to handle textile properties, in which the manipulation policy is conditioned on the cloth properties.

Task-wise, IL methods have been applied to cloth smoothing~\cite{seita2021learning, ganapathi2021learning, seita2020deep},  folding~\cite{salhotra2022learning, weng2022fabricflownet, lee2022learning,  ganapathi2021learning, xue2023unifolding}, twisting~\cite{jia2019cloth}, bed-making~\cite{seita2019deep}, and dressing~\cite{joshi2019framework}. Again, most works have only learned a single policy for one task~\cite{salhotra2022learning, lee2022learning, seita2020deep, jia2019cloth}, with only a few that either learn a single backbone and different output branches on the same backbone~\cite{xue2023unifolding}, or a shared correspondence~\cite{ganapathi2021learning} for different tasks.
% \dbm{Missing discussion on adaptation, if it has been evaluated or not.}

\subsection{Heuristic-based Manipulation}

The last discussed approach in textile manipulation is based on human-designed heuristic rules instead of learning the manipulation strategies from data. The heuristic rules vary depending on the target manipulation task. 
For example, for cloth smoothing, one commonly used rule is to detect the wrinkles of the cloth, and the manipulation action pulls the cloth in the perpendicular direction to the detected wrinkle direction~\cite{seita2019deep}. %The wrinkles can be detected using computer vision algorithms such as edge detection or by matching the contour of the manipulated garment to a template garment of the same category~\cite{doumanoglou2016folding}. 
For grasping, one rule is to use the %also to detect the wrinkle of the garment when lying on a flat surface, whose 
position and orientation of a wrinkle % can be used 
to compute the target location and orientation of the gripper~\cite{doumanoglou2016folding}. Another heuristic for a cloth article hanging in the air is to grasp key points that are identified based on the border geometry, e.g., corners for a towel or sleeves for a t-shirt~\cite{maitin2010cloth}, and use force sensors to trace the edge and get to the opposite corner~\cite{proesmans_2023_unfoldir}. %These border points can be estimated using discontinuities in the depth image.
For folding,  %one heuristic is to fit a polygon to the contour of the garment~\cite{miller2012geometric} and match it with a template garment of the same category. The folding plan can then be derived from the matched vertices~\cite{doumanoglou2016folding}. 
%Given a folding plan, 
a heuristic folding motion can be achieved by leveraging gravity and moving the cloth such that the moved part is always vertical, which is called a ``g-fold''~\cite{miller2012geometric}. 
For assistive dressing, a common heuristic solution is to move the grasped garment along the forward direction of the human limb~\cite{erickson2018deep}. 
%For a entire manipulation pipeline that involves multiple stages, e.g., first grasping and then folding, 
%Manually designed state machines are also frequently used to describe the whole manipulation procedure that involves multiple stages, e.g., first grasping and then folding, often taking into consideration error recovery~\cite{maitin2010cloth}. 

The generalization ability of such heuristically defined rules varies based on the target manipulation task and the assumptions made. 
Usually, these methods generalize well under the settings where the assumptions are satisfied and tend to have limited generalization when the assumptions are broken.  
Due to the diverse properties and configurations a textile can have, it can be hard to design a heuristic rule that would generalize to every situation one might encounter when manipulating textiles. 

\section{BENCHMARKS AND DATASETS}
\label{sec:resources}

%\todoinb{Irene- Complete subsections extending descriptions of citations}
%\todoinv{Michael, Alberta, Yufei}
%\todoinor{Irene, Alberta- Discuss during ICRA the intended outcome}
%\todoinor{Irene - Iterate over feedback.}
%\todoinor{Irene, Júlia- Shrink to 2 pages.}

% \al{The purpose of this section is twofold: 1) to understand what benchmarks and datasets are used to tackle the problem of adaptation to variations of physical properties, 2) to cover limitations of current benchmarks and datasets.}

% \begin{itemize}
%     \item Intro about benchmarking: Why are they important? How can they help to understand and validate the performance of the methods? Challenge: intertwinement between the different aspects of robotics (adressed in the sections of this paper). 
%     \item Intro to datasets: Objects are a crucial part in benchmarking
%     \item Benchmarks section: this will have the same structure as the previous ones (list of benchmarking papers (very few) and to what extend do they cover variations of props.
%     \item Datasets section: How papers that present datasets/object sets want to solve this issue. 
% \end{itemize}

%Past and current literature of robotic manipulation presents a wide variety in methodologies, experimental evaluation, robotics systems and objects used. Such differences, added to the lack of reproducible test conditions, usage of the same set of metrics and standardized objects makes difficult to provide a fair performance comparison. For this reason,
%In the last decade, benchmarking robotic manipulation has foster progress in this field. 

Benchmarking robotic manipulation plays a crucial role in understanding methods' capabilities and limitations, enabling standardized comparison. %This includes the verification of the adaptability and generalization of a proposed system through the introduction standardized scenes, measures and objects. %One way to indirectly validate this generalization is through the use of sets of objects covering a variety of properties.
%and the use of diverse objects.
Particularly benchmarking of cloth manipulation is limited due to the absence of objective and consistent evaluation processes and the limited research on how different textiles influence the performance of a method.

In the following subsections, we will discuss the two main tools for benchmarking: datasets and benchmarks. 
Datasets can demonstrate the adaptability of a method to new data. However, when physical interaction is inherent to the task, the generalization of a method needs to be demonstrated through a benchmarked experimental validation, including standardization of the objects used.
 
%However, due to the lack of an objective and consistent evaluation process, benchmarking is challenging in robotic manipulation in general, but specially limited for cloth manipualtion, due to the limited research on factors such as the effect of different textile materials, features and structures influence the performance of a method. Having such a quantitative comparison of performance will be of wide reaching benefit for this objective.

%\al{I would add here a brief description of what the next two sections will cover, related to the challenges of benchmarking introduced in the previous paragraph. In particular, I think it would be beneficial to distinguish between benchmarks and datasets.}
%Benchmarks enable the evaluation of robots in performing tasks at a mechanism, algorithmic, or systems level.

\subsection{Benchmarks}

Most benchmarks tackling cloth objects evaluate manipulation in simulated environments, %Simulation benchmarks have the advantage that they can automatically generate many different conditions, enabling the evaluation of generalization. For instance, SoftGym \cite{lin2021softgym} is a benchmark for deep reinforcement learning of deformable object manipulation that provides a set of simulated standardized environments and includes a set of cloth manipulation tasks. Assistive Gym \cite{erickson2020assistivegym} is a physics simulation framework for assistive tasks that includes an assistive dressing task with a hospital gown. 
since they have the advantage of automatically generating various conditions to evaluate generalization. For instance, SoftGym \cite{lin2021softgym} for deformable object manipulation and Assistive Gym \cite{erickson2020assistivegym} for assistive tasks, are simulation benchmarks for reinforcement learning that provides a set of simulated standardized environments.
 %For instance, DexArt \cite{bao2023dexart} evaluates the policies’ generality and robustness, applying them to unseen objects and changing the camera's viewpoint, although for rigid objects. 
However, due to the sim2real gap, it is necessary to have benchmarks for evaluating real-world applications with physical objects. Garcia-Camacho et al. \cite{garcia-camacho2020benchmarking} proposes benchmarks with real executions for three cloth manipulation tasks including protocols, qualitative evaluation metrics, and several complexity levels based on the initial state of the cloth. Alternatively, Clark et al. \cite{clark2023household} proposes four benchmark tasks for evaluating the performance of end-effectors in grasping clothing items, along with protocols to normalize crumpled configurations and metrics. %, and presents results for four off-the-shelf grippers. %In both cases, the authors present a small set of clothing items, not covering the variety described in Section \ref{sec:background}.

%Both works agree that an important aspect of benchmarks is the standardization of the objects. In addition, 
An effective way to measure the generalization of a method is through the use of a wide number of objects with varied properties. As it was done for rigid objects with the now widespread YCB object set \cite{calli2015ycb}, an extension focusing on textile objects is provided in \cite{garcia-camacho2022hcos}, which was distributed among the participants of the cloth manipulation and perception competition \cite{garcia-camacho2022competition}. %organized in IROS'22 and ICRA'23 to extend its usage, a requirement for a benchmark to be useful.  
It includes a wide variety of cloth household objects with benchmarking guidelines for its use. %This object set 
However, an important issue in defining standardized textile object sets is stock continuity, preventing the maintenance of the same objects for extended time periods. To solve this issue, a method for building comparable textile object sets across different publications has been proposed in \cite{garcia-camacho2024standardization}. 
It proposes a framework to characterize textile objects, enabling the quantification of variability on the textile properties listed in Section \ref{sec:background} of a given set of objects. 
With this characterization, the generalization of a method can be quantified based on the amount of variation that the objects offer. This idea goes in line with the one proposed for rigid objects in \cite{pumacay2024colosseum}, where generalization is measured similarly but also adds other aspects of variability in background, table color, etc.% Variability in the starting configurations and scene properties of color and light should also be considered for cloth manipulation.

\subsection{Datasets}

Unlike benchmarks, datasets serve the purpose of providing data for designing or learning a task. The extent of generalization depends on the variation covered by the dataset. Datasets are often task-specific, ensuring that the data is relevant both for training a system and evaluating its performance. In cloth manipulation, the most common datasets include simulated 3D models \cite{zhou2023clothesnet, bertiche2020cloth3d}, RGB images \cite{liu2016deepfashion, bednarik2018learning, Gustavsson2022Cloth} or depth \cite{avigal2022speedfolding,verleysen2020video}.

The largest existing datasets are for classic vision problems like cloth classification and landmark point detection, with datasets such as ClothesNet~\cite{zhou2023clothesnet} in simulation or DeepFashion~\cite{liu2016deepfashion} with real images. Datasets for segmentation of people wearing clothes % worn by humans, 
include CLOTH3D \cite{bertiche2020cloth3d} in simulation or \cite{zhao2018understanding} with real images. Surface reconstruction is another classic perception problem applied to cloth \cite{bednarik2018learning}. For cloth classification or landmark detection, images contain annotations of cloth type and location of landmarks, but usually, images come from the fashion industry, and so clothes are either flatted, hung on hangers, or worn by humans. For surface reconstruction, annotations need to have realistic images with the real mesh of the object, and therefore, existing datasets are all in simulation and subject to the sim2real gap discussed in Section~\ref{sec::model_gap}. 

For robotics, cloth classification and landmark point detection are also important, but they need to be identified during the stages of manipulation where clothes are in very complex configurations. Indeed, cloth classification is required but from crumpled states  \cite{sun2017single}, or when grasped by one point \cite{mariolis2015pose}.  Other features that need to be identified are corners and edges \cite{thananjeyan2022all} or wrinkles \cite{wagner2013ctu}. The dataset \cite{Gustavsson2022Cloth} is an expansion from DeepFashion~\cite{liu2016deepfashion} to adapt it to robotic manipulation.

One of the challenges for datasets in cloth manipulation is labelling the ground truth of the deformation state in real images. Only a few datasets exist with depth or point clouds \cite{verleysen2020video} and even less with labels of what points correspond to corners or edges during a manipulation \cite{schulman2013tracking}. So far, to encode interactions, some datasets use RGB-D images where the action is annotated as a pick-up pixel point and a direction of motion in the image \cite{avigal2022speedfolding}, with the limitation of only representing close-to-planar configurations. More complex actions appeared lately with Visual Language Models where a sequence of images is linked to a sequence of positions of the end-effector \cite{chi2024universal}, with corresponding datasets.

Datasets rarely annotate the variability in the mechanical properties mentioned in Section \ref{sec:background}, and some of the physical properties are covered depending on the requirements of the trained system. The main issue is the difficulty in labeling the deformation ground truth from images due to the severe self-occlusions, while datasets in simulation are only partly useful due to the sim2real gap.

%https://docs.google.com/spreadsheets/d/1rPBD77tk60AEIGZrGSODwyyzs5FgCU9Uz3h-3_t2A9g/edit#gid=0
\section{APPLICATIONS AREAS}
\label{sec:application}

%\todoinb{Marco- Integrate citations (DONE)}
% \todoinb{Marco- Iterate fashion industry ction (DONE)}
%\todoinb{Marco- Fill in missing applications (DONE)}
%\todoinv{Hang, Irene, Alberta}
% \todoinor{Marco - Merge applications that have similar requirements}
% \todoinor{Marco - Iterate over feedback}
% \todoinor{Marco, Zackory - Shrink to 2 pages}
% \todoinor{Marco - Add citation to the table}
% \todoinor{Marco - Add reference to the table and some discussion over how much some tasks are addressed or not.}

% The purpose of the section: Provide an overview of the current tasks considered for manipulation of deformable objects and how these would be in a scenario where robots can successfully adapt to different variations of these objects.

% {\color{blue} Considering Physical properties variations in deformable textile objects has a wide range of applications.
% in general, the potential is greatest if either the object properties used in a particular task vary considerably, like in dressing tasks, and/or when the task specifically relies on the exploitation of certain properties i.e. elasticity of a bedsheet for bed making or the friction of a cloth used for wiping a surface clean.
% In this section, we will shortly discuss relevant current and potential applications in three settings, Health-care, household, and the Fashion industry.
% }

We outline the applications and challenges of manipulating deformable textile objects in scenarios requiring robots to adapt and generalize %\dbm{and generalize? we might be using it interchangeably} 
to their varying properties. Understanding the physical properties of textiles is crucial for a wide range of tasks in diverse sectors such as household chores, healthcare, and the textile industry. Table \ref{tab:task_categories} provides a (non-exhaustive) overview of work categories in these sectors, organized by the frequency at which they are addressed in the literature. Tasks such as folding, smoothing, and dressing receive frequent attention from the community, whereas tasks like buttoning, dyeing, and washing remain rather underexplored. In what follows, we will discuss in detail these tasks and the requirements concerning variations of textile properties.

% {\color{blue} for a wide range of tasks in diverse sectors such as household, Healthcare, and tasks needed in the fashion industry. In table \ref{tab:task_categories}, we have a (non-exhaustive) overview of work categories in the different sectors. Furthermore, we organized the addressed tasks by the frequency at which they are addressed in the literature. As the table shows, tasks such as Folding, smoothing, and dressing receive frequent attention from the community, while tasks such as Buttoning, dyeing, and washing remain rather underexplored. }

% {\color{red}
% for applications such as robot-assisted dressing, which shows significant object variation, or bed-making and surface wiping, which depend on specific properties like elasticity and friction. 
% % \al{Add a reference to the table and introduce the notion of what has been addressed and what not so far?}
% In this section, we explore these and other current applications and their challenges across healthcare, household chores, and the fashion industry. However, as illustrated in Table~\ref{tab:task_categories}, the distribution of research contributions across these applications is not uniform. The literature frequently addresses certain tasks while others remain under-explored. This highlights the disparities in focus and indicates which tasks require further attention to achieve a more balanced advancement in the field.
% }

\begin{table}[b]
\centering
\caption{Overview of Variation of Tasks addressed by the community and their frequency. }
\label{tab:task_categories}
\begin{tabular}{|p{3cm}|p{3cm}|p{3cm}|p{3cm}|}
\hline
\rowcolor[gray]{0.9} \text{Frequency} & \textbf{Household } & \textbf{Healthcare } & \textbf{Textile Industry } \\ \hline
\textbf{Frequent (4+)} & 
Folding~\cite{salhotra2022learning, doumanoglou2016folding, weng2022fabricflownet,   avigal2022speedfolding,ganapathi2021learning,  yang2016repeatable, koganti2014real} \newline Smoothing~\cite{hoque2022visuospatial, wu2019learning, ha2022flingbot, ganapathi2021learning, seita2019deep} \newline Ironing~\cite{li2016multisensor, dai2004trajectory, estevez2020enabling, estevez2017robotic, estevez2017ironing} & 
Dressing~\cite{ wang2023one, erickson2018deep,  koganti2014real, li2021provably, zhang2019probabilistic, yamazaki2014bottom, canal2019adapting,kapusta2019personalized, erickson2019multidimensional} 
% \newline Bathing~\cite{erickson2019multidimensional, madan2024rabbit, zlatintsi2020support, liu2024skingrip}
&  \\ \hline

\textbf{Rare(2-3)} & 
Hanging~\cite{matas2018sim, antonova2021sequential} \newline Sorting~\cite{sun2017single, kampouris2016multi} \newline Wiping~\cite{leidner2019cognition, dometios2018vision}  & 
Bedding~\cite{puthuveetil2022bodies, yang2016repeatable, goldberg2022deep} \newline Bed-making~\cite{seita2019deep, yang2016repeatable, goldberg2022deep}\newline Bandaging~\cite{longhini2023edonet, li2022method} & 
Recycling~\cite{kampouris2016multi, damayanti2021possibility}\\ \hline

\textbf{Unaddressed (0-1)} & 
Storing & 
Buttoning~\cite{fujii2022buttoning} & 
Manifacturing~\cite{gries2018application} \newline  Dyeing~\cite{papoutsidakis2019advanced}  \newline  Quality control \newline Coloring  \newline Washing \\ \hline
\end{tabular}
\end{table}

\subsection{Healthcare}

% \al{Q: What are the most common tasks performed? Why don´t we scale up? What are the bottlenecks? What could or should be done in the future? To what degrees do different properties matter the most?}
% \begin{itemize}
%     \item Dressing
%     \item Body covering and uncovering
% \end{itemize}

% {\color{blue}
% One area that has received considered attention is assisted dressing {\color{red} ZACKORY WORKS}

% further assistive tasks have the potential to benefit form the varied properties setting such as the autonomous covering/uncovering of a bedridden patient. In {\color{red} ZACKORYS GROUP WORK}

% Another health application is the changing of bandages, in this setting the {\color{red} EDO MOTIVATION TASK HERE}
% }

As populations age worldwide~\cite{world2015world}, there is a growing opportunity for robotic systems that provide physical assistance with activities of daily living (ADLs) and healthcare tasks. Different assistive tasks involve manipulating textile objects, such as in robot-assisted dressing~\cite{koganti2014real}, bedding~\cite{goldberg2022deep}, bathing~\cite{erickson2019multidimensional}, and medical care~\cite{li2022method}. This section examines significant advancements in healthcare robotics, particularly robot-assisted dressing, bathing and bedding, bed-making, and medical care, highlighting opportunities and challenges in real-world applications.

Robot-assisted dressing 
%and bathing 
is crucial for individuals with upper or lower extremity mobility impairments, requiring careful manipulation of deformable objects such as garments.
% and sponges.
This task necessitates the ability to adapt to various object variations while ensuring the safety and comfort of the patient. Research in robot-assisted dressing has led to the development of robots that can assist with putting on shirts~\cite{ koganti2014real, li2021provably, zhang2019probabilistic}, pants~\cite{yamazaki2014bottom}, and footwear~\cite{canal2019adapting}, with challenges including ensuring physical safety, accurately modeling human-robot interactions during garment occlusions~\cite{erickson2018deep, kapusta2019personalized, erickson2019multidimensional}, and generalizing to different garments~\cite{wang2023one}.
%Meanwhile, robot-assisted bathing~\cite{madan2024rabbit, zlatintsi2020support, liu2024skingrip} focuses on safely manipulating bathing objects, requiring precise force application and understanding the objects' physical properties to avoid harming patients. This highlights similar challenges to robot-assisted dressing, emphasizing the need for adaptive methods that ensure patient safety and comfort.

% \subsubsection{Bed-making and robot-assisted bedding}
Bed-making and blanket manipulation represent significant opportunities for cloth handling in caregiving involving large, deformable textiles. Research has led to robotic systems capable of grasping and smoothing fitted sheets~\cite{goldberg2022deep}, folding and arranging blankets and towels~\cite{avigal2022speedfolding, yang2016repeatable}. Key to effective bed-making is leveraging physical properties like elasticity of bedsheets during robotic manipulation~\cite{seita2019deep}, also for tasks like autonomously covering and uncovering a person in bed~\cite{puthuveetil2022bodies}.

% \subsubsection{Medical care}
In daily medical care, robotics research has introduced advances in handling soft materials such as gauze for bandaging~\cite{li2022method} or adult diapers~\cite{baek2023smart}. These tasks, involving physical contact with the human body, underscore the importance of incorporating various sensory modalities and control techniques for effective manipulation of soft materials~\cite{longhini2023edonet}.

\subsection{Household chores}

% \begin{itemize}
%     \item Folding
%     \item Flattening
%     \item Ironing
%     \item Laundry, sorting
%     \item Wipe 
%     \item Hanging
%     \item Bed making
% \end{itemize}

% {\color{blue}
% Our homes are full of cloth-like deformable objects and at the current forefront of robots that venture into completely unstructured spaces.
% One of the most anticipated stacks of tasks which still has to be realized to a sufficient extend, is doing laundry.
% Laundry can be seen as a composite task consisting of a number of subtasks, which starts with a pile of used, potentially dirty cloths

% }

Several instrumental activities of daily living (iADLs), such as laundry, cleaning with towels (fabric or paper), and hanging clothes, require dexterous textile manipulation. Cloth-like objects are ubiquitous in unstructured domestic environments and pose significant challenges to fully automating these activities. This section outlines methodologies and challenges in cloth sorting, smoothing, ironing, folding, hanging, and wiping tasks.

% \subsubsection{Sorting}
Cloth sorting involves categorizing garments and textiles by attributes such as item class~\cite{sun2017single}, fabric type, construction, color, and quality, generally before washing or recycling. Accurate perception and classification of these variations can enhance sorting efficiency, enabling generalization across different textile batches.% for more effective sorting\dbm{this feels repetitive: "can enhance sorting efficiency ... for more effective sorting"}.
While recent methods have incorporated material identification through tactile feedback~\cite{kampouris2016multi}, few integrate physical interaction to accurately discern mechanical properties.

Robotic tasks such as smoothing, ironing, and wiping involve manipulating fabrics under varying physical conditions, which are primarily influenced by properties like friction and elasticity. Smoothing~\cite{hoque2022visuospatial, wu2019learning, ha2022flingbot, ganapathi2021learning, seita2019deep}, typically performed before folding or wiping, necessitates to account for varying friction as fabrics transition from crumpled to flat states. Ironing further requires adjustments for temperature variations that affect the fabric's physical properties. Current methods mainly focus on ironing individual garments~\cite{li2016multisensor, dai2004trajectory, estevez2020enabling, estevez2017robotic, estevez2017ironing} and often fail to generalize across different fabric types. Wiping involves tackling the challenges posed by variable surface friction, which can be influenced by the presence of dust or liquids on different surfaces~\cite{leidner2019cognition, dometios2018vision}. Each task demands adaptive strategies to cope with the dynamic nature of fabric properties, highlighting the need for methods that can accommodate these variations, as discussed in Section ~\ref{sec:manipulation}.

% \subsubsection{Folding}
Significant research in robotic manipulation of cloth has focused on robotic folding with practical applications in healthcare and domestic settings~\cite{ salhotra2022learning,doumanoglou2016folding,  weng2022fabricflownet, avigal2022speedfolding, ganapathi2021learning,  yang2016repeatable,koganti2014real}. However, challenges such as variability in garment properties (weight, friction, and shape) affect both quasi-static and dynamic manipulation~\cite{blanco2024benchmarking}, impacting generalization. Notably, material stiffness, as reflected in bending coefficients, significantly influences each fold, potentially accumulating errors and altering outcomes in the folding process~\cite{longhini2024adafold}.
%Newer methods also introduce dynamic actions like flinging to obviate the need for garment smoothing~\cite{ha2022flingbot, avigal2022speedfolding}.

% \subsubsection{Hanging}
Cloth hanging involves finding a stable configuration for a garment on a hanger~\cite{matas2018sim} by identifying features like holes and loops~\cite{antonova2021sequential}. However, the impact of the physical properties of the garment on the deformation of these features remains largely underexplored.

\subsection{Textile Industry}
The textile industry, encompassing sectors such as fashion, automotive, and construction, presents numerous opportunities for autonomous robots with adaptive capabilities. This subsection focuses on specific applications such as manufacturing, dyeing, and recycling, directly linked to adaptability in handling, perceiving, and quality management of textiles. Given the extensive previous discussion on sorting, smoothing, ironing, and folding, this section will concentrate on these industry-specific applications.

% \subsubsection{Manifacturing}
In cloth manufacturing, robots are increasingly integrated into the cutting and sewing stages of garment production~\cite{gries2018application}, with potential advancements enabling them to detect variations in fabric properties like thickness and stretchability. This ability could allow precisely adjusting techniques for each material, significantly reducing waste and enhancing resource efficiency, thus improving sustainability and garment quality.

% \subsubsection{Dyeing}
The automation of dyeing processes significantly enhances sustainability by reducing dye and water usage~\cite{papoutsidakis2019advanced}. Currently, robots in dyeing processes are mainly used for loading and unloading yarn bobbins. Augmenting these robots to recognize variations in garment properties could optimize dye application, tailoring it to the specific needs of each garment.

% \subsubsection{Quality control:}

% Quality control ensures that textiles meet specific standards before they proceed further in the production chain or reach the consumer. Adaptability in quality control could involve the use of automated systems that adjust inspection parameters based on the type of fabric being examined. This could include adaptations for different weaves, thicknesses, or finishes to identify defects or deviations from expected quality standards accurately.

% \subsubsection{Recycling}
Finally, recycling is crucial for sustainability~\cite{damayanti2021possibility}. Similar to sorting, enhanced interactive perception of fabric types, construction methods, and material conditions can improve this process by accurately identifying textiles suitable for recycling~\cite{kampouris2016multi}.

% \item \subsubsection{Washing:} Adaptive technologies in washing processes can optimize water usage, detergent amounts, and washing cycles based on the load’s characteristics, such as weight, material type, and garment condition. This not only improves the efficiency and sustainability of washing processes but also extends the lifespan of garments by using gentler or more appropriate wash settings.

\section{DISCUSSION AND FUTURE PERSPECTIVES}
\label{sec:conclusions}

In previous sections, we examined modeling, perception and manipulation separately. However, there is significant interplay among these domains, impacting generalization and adaptation capabilities. % Here, we review some challenges faced by these components individually and how their integration can be beneficial for generalization.
Offloading computational effort to modeling allows for model-based optimization of manipulation trajectories, reducing the burden on control and perception~\cite{li2015folding}. However, real-world applications need perception techniques to align the parameters of the model to the real-world object, making perception crucial for generalization~\cite{longhini2024adafold, longhini2023edonet}. While model-free learning techniques like RL and IL learn end-to-end from raw data~\cite{hietala2022learning}, reducing perception needs, they sacrifice sample efficiency and generalizability to novel variations of the environment and tasks. Nonetheless, not all components of manipulation tasks need end-to-end learning; perception modules can simplify sub-tasks like flattening wrinkles~\cite{sun2014heuristic}, grasp point detection~\cite{corona2018active}, and folding plans, bypassing the need for extensive learning.  %While relying solely on perception and heuristics has been explored, it also limits generalizability, justifying learning policies that can address tasks where heuristics fail. 
As an example,  perception can determine grasp points, how to reach the grasping point can be resolved with a standard planning algorithm, and how to optimize the manipulation once the cloth is grasped can be learned and performed by an end-to-end policy. Exploring when to switch between learning and heuristics is a promising research direction. This is particularly relevant with the advent of foundation models capable of reasoning about semantics, sequential tasks, and adapting to rules and human preferences.

In the remainder of the section, we further identify open problems and detail future research directions and grand challenges to foster the development of perception and manipulation skills that generalize to variations of textile properties and manipulation tasks.

\subsection{Open Problems}

% To discuss the generalization and adaptation of current methods for cloth manipulation, we provided an overview of fundamental textile properties that contribute to the variations of real-world textiles, highlighting the complexity of understanding their role in manipulation due to the intertwined role of each of them. However, the influence of each textile property in robotic manipulation is still underexplored. Moreover, the extent to which explicit identification of each property is required remains unclear. 
% An open avenue for research is identifying a subset of pertinent features for different manipulation tasks, which can be used to determine the extent to which the textile properties need to be identified. 

To discuss generalization and adaptation in cloth manipulation, we reviewed fundamental textile properties and their complex, intertwined roles. However, the influence of each property in robotic manipulation is still underexplored, and the necessity of explicitly identifying each property remains unclear. An open avenue for research is identifying a subset of pertinent features for different manipulation tasks to determine the extent to which textile properties need to be identified.

% Adding to the complexity of identifying and defining the textile properties is the dynamic nature of these properties under varying conditions. Aging of the cloth, as well as wet, dirty, or dry conditions, can alter a textile's behavior, emphasizing the need for continuous sensing and adaptive perception systems in autonomous agents. Moreover, physical attributes like thickness, softness, and durability~\cite{yuan2018active} are typically evaluated alongside semantic descriptors, like smooth, absorbent, hairy, and slippery, that humans use to characterize textiles~\cite{chu2013using}. Particularly, with the advances in Large Language Models (LLMs) there is an emerging potential to bridge these descriptive terms with a physical understanding of textile properties~\cite{yu2024octopi}.

Adding to the complexity of defining textile properties is their dynamic nature under varying conditions. Aging, wetness, dirt, and dryness can alter a textile's behavior, necessitating continuous sensing and adaptive perception in autonomous agents. Physical attributes like thickness, softness, and durability~\cite{yuan2018active} are often evaluated alongside semantic descriptors like smooth, absorbent, hairy, and slippery~\cite{chu2013using}. Advances in Large Language Models (LLMs) present potential to bridge these descriptors with a physical understanding of textile properties~\cite{yu2024octopi}.

One key aspect for manipulating a variety of textiles and addressing the variation of properties is perception. While interactive perception with different types of sensors and exploratory actions can enhance adaptability and reduce uncertainty, it remains underexplored. Multimodal sensing emerges as a pivotal strategy in this context, integrating various sensory inputs like tactile, visual, and auditory data to provide a more holistic understanding of textile properties.  This approach holds significant potential, as it allows robots to leverage multiple temporal information sources, compensating for the limitations of individual sensors.

% David: I liked the "A significant observation" :)
% A significant observation of our review of manipulation techniques is the limited amount of works demonstrating effective generalization to a wide range of deformable objects, with few exploring adaptation methods.  Most current efforts utilize domain randomization in simulations to enhance sim2real transfer. While promising, these methods often overlook the potential of various domain adaptation techniques that could allow robots to dynamically adjust their manipulation strategies based on real-time environmental feedback and changes in textile properties.
% Similarly, the use of state-of-the-art models such as diffusion models, in combination with imitation learning techniques, has predominantly focused on rigid objects due to the complexity of state estimation and high-dimensional state spaces of textiles. This leaves open questions about their applicability to deformable objects and the challenges of deploying these methods without assuming rigidity. 

A significant observation of our review of manipulation techniques is the limited amount of work demonstrating effective generalization to a wide range of deformable objects, with few exploring adaptation methods. Most efforts use domain randomization in simulations for sim2real transfer, often overlooking domain adaptation techniques that allow dynamic adjustment based on real-time feedback due to perceptual challenges. 
Similarly, state-of-the-art models, such as diffusion models, in combination with imitation learning techniques, remain underexplored. This leaves open questions about their applicability to deformable objects and the challenges that might arise due to the complexity of state estimation and high-dimensional state spaces of textiles.

% Building on the necessity for improved generalization in manipulation techniques, benchmarking, standardization, and datasets emerge as essential yet challenging areas. The diversity of robotic embodiments, sensor configurations, and test sets, complicates the comprehensive comparison of approaches and algorithms. Specifically, the development of universally applicable testing datasets is hindered by issues like stock availability and the difficulties associated with measuring properties accurately. Although there has been some progress towards creating standardized test sets and benchmarks, these continue to be significant open problems within the field. Additionally, a major challenge in dataset availability and scalability is accurately labeling the deformable object dataset, which limits their widespread distribution and application.

Building on the necessity for improved generalization in manipulation techniques, benchmarking, standardization, and datasets emerge as essential yet challenging areas. The diversity in robotic embodiments, sensor configurations, and test sets complicates comprehensive comparisons. Developing universally applicable testing datasets faces challenges like stock availability and accurate property measurement. Although progress has been made, standardizing test sets and benchmarks remains a significant problem. Additionally, a major challenge in dataset availability and scalability is accurately labeling deformable object datasets, limiting their widespread distribution and application.

\subsection{Grand Challenges}
To push the boundaries of robotic manipulation of textiles, we identify the following critical challenges that may have major breakthroughs in the coming years:

% \begin{itemize}
% Perception of new textiles: Develop methods to accurately estimate key properties of new textiles. This requires not only understanding the properties of interest but also employing combinations of sensing and exploratory procedures designed to identify these properties effectively.

% \textbf{Perception of properties of novel objects:} Develop advanced methods for accurately estimating key physical and semantic properties of new textiles. This challenge requires integrating sophisticated sensing technologies and exploratory procedures to identify these properties effectively. Achieving this would enhance robotic manipulation, enabling robots to adaptively handle a diverse range of textiles.

\textbf{(i) Perception of properties of novel objects} for accurately estimating key physical and semantic properties of new textiles, enabling robots to reduce the uncertainties about the environment and adaptively handle a diverse range of textiles.
% \item Multi-modal, Multi-task, Multi-embodiment Generalist Agent: Create an agent capable of addressing multiple tasks in unknown environments, particularly those involving the manipulation of deformable objects and performing assistive tasks with humans in the loop. Challenges include learning to autonomously adapt to complex situations and acquiring skills from minimal human demonstrations.

% \textbf{Multi-modal, Multi-task, Multi-embodiment Generalist Agent: } Develop an agent capable of addressing multiple tasks in unknown environments, especially those involving the manipulation of deformable objects and assistive tasks with humans in the loop. This requires the agent to autonomously adapt to complex situations and acquire skills from minimal human demonstrations, leveraging multi-modal sensory inputs and versatile embodiment strategies. Achieving this would significantly advance robotic flexibility and effectiveness in diverse real-world applications.

\textbf{(ii) Adaptive multi-task and multi-modal agent} capable of autonomously adapting to complex situations, performing long-horizon tasks, and using multi-modal sensory inputs to navigate uncertainties, enabling seamless integration into homes, industries, and healthcare facilities.

% \item Datasets and Benchmarking: Address the difficulties in creating comprehensive datasets due to the high-dimensional state space and degrees of freedom. A well-annotated dataset encompassing the 2D state across various dynamic interactions and labeled with relevant features and properties would greatly benefit the community. Additionally, designing experimental benchmarking procedures for textile objects that can be standardized across institutions, considering stock availability and object variations, is crucial for progress.
% \end{itemize}

% \textbf{Datasets and Benchmarking:} Address the challenges in creating comprehensive datasets for high-dimensional state spaces and degrees of freedom in robotic manipulation. Develop a well-annotated dataset that captures 2D or 3D states across diverse, dynamic interactions labeled with relevant features and properties. Additionally, design standardized experimental benchmarking procedures for textile objects, considering stock availability and object variations, to facilitate consistent progress across research institutions. Achieving this will significantly enhance the community's ability to develop and compare advanced robotic systems.

\textbf{(iii) Novel datasets and benchmarks}  capturing real-world variations in object properties, physical interactions, and manipulation tasks to facilitate standardized benchmarking and enable consistent comparisons of robotic systems across research institutions.

The field of robotic manipulation of textiles is extensive and includes numerous important technological areas not mentioned here. Thus, the list provided is not exhaustive. The three grand challenges identified — perception of properties of novel objects, adaptive multi-task and multi-modal agents, and novel datasets and benchmarks — represent critical areas that have the potential to drive major advancements. These challenges encompass core perceptual technologies, general and adaptive capabilities, and standardized evaluation methods, aiming to enhance the flexibility and effectiveness of robotic systems in diverse real-world applications.

% Grand challenges to discuss
% \begin{itemize}
%     \item Characterization and Perception of Textile Properties: there are different levels of abstractions of properties that we do not consider but that are relevant in real-world applications. Their relevance and how they are perceived contain a lot of potential future directions to explore.
%     \item Manipulation techniques: there is a lot for rigid in terms of diffusion as well as adaptation techniques, but not for deformable. \al{What about modeling? Analytical models will never be good, but we could explore better ways to exploit them as priors, and this direction has definitely gained some interest recently, also thanks to differentiable simulators.}
%     \item Variations of tasks: some tasks are not addressed.
%     \item Benchmarking, standardization.
% \end{itemize}

% Examples provided by default in the review template. To be removed
% \include{examples}

\section*{AUTHORS CONTRIBUTIONS}

A.L. conceived the idea and the structure of the review, authored sections 1,2,4,8, coauthored sections 3,5, revised the full manuscript, created Figures 1 and 3, contributed to the creation of Tables 1 and 2, and coordinated the overall preparation and revision of the manuscript. 
Y.W. authored section 5, revised the full manuscript, and contributed to the creation of Table 1.
I.G.-C. authored section 6, coauthored section 2, revised the full manuscript, and created Figure 2.
D.B.-M. coauthored section 5, revised the full manuscript, and contributed to the creation of Table 1.
M.M. authored section 7, revised the full manuscript, and contributed to the creation of Table 2.
M.W. assisted with the delineation of the structure of the review, contributed to section 3, and revised the full manuscript.
H.Y. coauthored section 3 and revised the full manuscript.
Z.E. coauthored section 7 and revised the full manuscript.
D.H. coauthored section 5 and revised the full manuscript.
J.B. coauthored section 6 and revised the full manuscript.
G.A. and D.K. revised the entire manuscript.

\section*{ACKNOWLEDGMENT}
This work has been supported by the European Research Council (ERC-BIRD); the European Union’s Horizon Europe Programme through projects SoftEnable (HORIZON-CL4-2021-DIGITAL-EMERGING-01-101070600) and IRE (HORIZON-CL4-2023-DIGITAL-EMERGING-01-101135082); the Swedish Research Council, the Knut and Alice Wallenberg Foundation, and the National Science Foundation under NSF CAREER grant number IIS-2046491; the project ROB-IN PLEC2021-007859 funded by MCIN/ AEI /10.13039/501100011033 and by the European Union NextGenerationEU/PRTR.

\bibliography{references} 
% We will have to switch to ar-style but I have experienced problem with citet so far, so for now I changed style.
 \bibliographystyle{ar-style3} 

\end{document}